\documentclass{article}
\usepackage{arxiv}

\usepackage[utf8]{inputenc} 
\usepackage[T1]{fontenc}    
\usepackage{hyperref}       
\usepackage{url}   
\usepackage{booktabs}       
\usepackage{amsfonts}       
\usepackage{nicefrac}       
\usepackage{microtype}      
\usepackage{lipsum}
\usepackage{fancyhdr} 
\usepackage{graphicx}  
\usepackage{lscape}
\usepackage{amsmath}
\usepackage[round]{natbib}
\usepackage{tabularx}
\usepackage{makecell}
\usepackage{todonotes}
\usepackage[font=small,skip=0pt]{subcaption}
\usepackage{caption}
\title{Leveraging Image-based Generative Adversarial Networks for Time Series Generation}

\author{
Justin Hellermann \\
\textbf{Stefan Lessmann}\\
\\
Humboldt University of Berlin \\
School of Business and Economics\\
Unter den Linden 6\\
10099 Berlin\\
\texttt{\{justin.hellermann, stefan.lessmann\}@hu-berlin.de} \\
}

\renewcommand{\shorttitle}{Leveraging Image-based Generative Adversarial Networks for Time Series Generation}

\hypersetup{
pdftitle={Leveraging Image-based Generative Adversarial Networks for Time Series Generation},
pdfsubject={},
pdfauthor={Justin ~Hellermann, Stefan ~Lessmann},
pdfkeywords={Intertemporal Return Plots, IRP, Generative Adversarial Networks, Time Series, Recurrence Plots},
}

\begin{document}
\maketitle
\begin{abstract}
Generative models for images have gained significant attention in computer vision and natural language processing due to their ability to generate realistic samples from complex data distributions. To leverage the advances of image-based generative models for the time series domain, we propose a two-dimensional image representation for time series, the Extended Intertemporal Return Plot (XIRP). Our approach captures the intertemporal time series dynamics in a scale-invariant and invertible way, reducing training time and improving sample quality. We benchmark synthetic XIRPs obtained by an off-the-shelf Wasserstein GAN with gradient penalty (WGAN-GP) to other image representations and models regarding similarity and predictive ability metrics. Our novel, validated image representation for time series consistently and significantly outperforms a state-of-the-art RNN-based generative model regarding predictive ability. Further, we introduce an improved stochastic inversion to substantially improve simulation quality regardless of the representation and provide the prospect of transfer potentials in other domains.
\end{abstract}

\section{Introduction}
\label{sec:intro}
Simulation of times series is a backbone in financial, economic, and engineering applications and serves the purpose of evaluating products or making probabilistic predictions. Financial institutions run complex Monte Carlo simulations to quantify risks, price financial products and commodities, or identify patterns for trading strategies \citep{drew2012}. The models often require large amounts of data to make accurate predictions, and real-world data may be scarce. Simulating time series data enables the creation of synthetic data that resembles real-world data for various purposes, including modelling, prediction, and analysis. In financial applications, synthetic time series facilitate the modelling of stock prices, interest rates, and other financial instruments to make informed predictions about future market trends and evaluate investment strategies' performance. In economic applications, synthetic time series data can support estimating variables such as GDP, inflation, and unemployment to strengthen economic predictions and evaluate the impact of shocks or policy changes. Also, in engineering applications, the simulation of time series data is used to model the performance of power grids, transportation networks, and communication systems. High-quality simulated data enables engineers to assess system performance and monitor the impact of design changes.

Simulation techniques often rely on parametric approaches to generate time series. In the first step, parametric models assume a reasonably similar data-generating process. Afterwards, the models estimate coefficients for a forecasting model that has been selected based on the assumption of the process. However, fitting a forecasting model across time is risky, especially since time series can visit different regimes with opposing characteristics. Despite the drawback that the number of regimes is often specified beforehand and excludes the possibility of other unobserved regimes in unseen data, financial applications frequently use regime-switching models \citep{HERRERA2018622,SCHARTH2009304}.
Once multiple time series are involved, training and calibrating the model becomes tedious since models rely on estimating correlations between many features, and regime switches result in non-stationary correlations over time \citep{SCHAFER20103856}. However, the higher the expectation of the model's accuracy and generalization, the more training data is required. Further, some behaviour is rare and may only occur scarcely but is highly important for the business context. A miscalibrated model due to the data shortage might have far-reaching consequences. Solutions to the problems may be a complexity reduction, sacrificing some forecast accuracy, or synthesizing additional data that can help to fit parameter models.

Creating artificial data that resembles real-world data is frequently used in computer vision to train models and algorithms. Synthetic data has several advantages over real-world data, such as being cost-effective, easier to obtain, and often free from privacy and ethical issues. Further, synthetic data allows balancing datasets, creating datasets with specific characteristics, and training models in cases where real-world data is scarce or unavailable. Recently, deep learning models augmented the set of models to generate new images that are similar in appearance to a given training set of images. The dominant classes of the models do not assume that the generated images come from a particular distribution, such as a Gaussian or a uniform distribution. Instead, they learn the underlying distribution of the training data via adversarial training of a generative and a discriminative network. They generate diverse and complex images similar to the training data and learn complex and multi-modal distributions.\\
Especially the work of \citet{karras_style-based_2019} has shown high-quality synthetic data of faces with disentangled, controllable features. While generative adversarial networks (GANs) have long been a quasi-standard for generative models, the new class of diffusion models has recently surpassed GANs in generating artificial images in a benchmark study \citep{NEURIPS2021_49ad23d1}. The current successes of text-to-image models impressively prove the capabilities of conditional image generation \citep{dalle2021} and can contribute to art, design, and content creation.
The image-based generative GAN compensates for the upper drawbacks of classic parametric models by not assuming the data-generating process, which allows incorporating non-stationarity easily and makes no assumptions on regimes or intertemporal relationships. A crucial step, however, is to find an efficient image representation for the time series to leverage existing image-based approaches and benefit from their modelling advantages and generative abilities.
Therefore, it needs to represent the time series in a two-dimensional space that preserves the characteristics of the time series and allows for direct inversion after the synthesis process.
\\
With this article, we aim to (1) propose a novel methodology named extended intertemporal return plot (XIRP) for time series capable of capturing the dynamics of time series, (2) validate our approach with simulated and real data, (3) perform a benchmark test to a state-of-the-art generative model, and (4) provide guidance for a stochastic inversion of the image representation to increase the predictive ability of the synthetic samples.\\
\\
The new image representation considers scale-invariant intertemporal relationships stored in the off-diagonal of the image while the diagonal contains the initial time series. So, it preserves both intertemporal relationships and raw time series values. This results in three significant advantages of the proposed approach:\\
\\
Firstly, the XIRP captures the intertemporal dynamics of a time series better than other representations. Motivated by principles from finance, it encodes the initial time series and its intertemporal returns in a single image, allowing for easy deterministic and stochastic inversion. 
\\
\\
Secondly, the proposed XIRP proves competitive and surpasses other image representations regarding predictive ability across a range of synthetic and real datasets.\\
\\
Thirdly, our approach allows for reducing the amount of complexity to a large extent.
Instead of relying on multiple steps to encode and decode, the model processes the XIRP directly and samples new image representations, which are then inverted to obtain a time series.\\
\\
Our results have multiple implications for research and practice. On the research side, our approach motivates the use of image-based representations for various advanced generative models. The XIRP is suitable for all types of image-based generative models, and its design extends the field of application beyond generation toward regression and classification tasks. 
The XIRP enables exploration, review, and improvement of engineering, medical, and energy applications that have so far relied on other image representations, e.g. for fault diagnosis \citep{cui_lightweight_2022}, biosignal classification \citep{DEPAULA2023119096} and simulation of scenarios for renewable energy generation \citep{wang_quantitative_2022}. 
Augmented data based on XIRPs, potentially improve any parametric and nonparametric model regarding generalization ability and overall performance. The results can be directly transferred  to make forecasting models more robust against adversarial attacks \cite{sarkar2020neural} and improve forecasting accuracy \citep{ELALEM2022,zhang_data_2022}. \newline
To our knowledge, this paper is the first to evaluate an invertible image representation of time series for generative models regarding similarity and predictive ability and compares the results to state-of-the-art benchmarks. We hope our contribution intensifies the research with newly developed image-based generative models and encourages practitioners to incorporate data augmentation for time series as a standard part of the model development workflow.

\section{Background}
\label{sec:background}
The chapter provides a comprehensive introduction to generative adversarial networks and time series images, which are the foundation of the paper. It also introduces concepts relevant to our research questions and hypotheses and aims to provide a thorough understanding of the context. Eventually, the discussion enables readers to appreciate the impact and contribution of our propositions in the areas of time series representations, generative models, and data augmentation.

\subsection{Generative Adversarial Networks}
From the generative aspect, traditional statistical methods and machine learning techniques often fail to capture the temporal dependencies, non-linear patterns, and multi-model distributions of time series data \citep{lipton_learning_2017}. GANs have proven effective in addressing these issues. They emerged as a powerful tool for learning the underlying data distribution and generating realistic synthetic samples. The generative concept relies on two neural networks, the generator $G$ and discriminator $D$, trained in an adversarial setting. The generator takes a noise vector sampled from a probability distribution as input and generates synthetic samples. In contrast, the discriminator receives real and synthetic samples and attempts to distinguish between them. The generator and discriminator networks are trained iteratively in a minimax game until the generator can produce synthetic samples indistinguishable from real data. The objective function of the adversarial game between $G$ and $D$ is given by
\begin{equation} \label{eq:vanillaGan}
\min_{G} \quad \max_{D}\quad \mathop{\mathbb{E}}_{x\sim \mathop{\mathbb{P}_r}}[log(D(x))]+\mathop{\mathbb{E}}_{\tilde{x}\sim \mathop{\mathbb{P}_g}}[log(1-D(\tilde{x}))]
\end{equation}
where the discriminator and generator try to maximize and minimize (3), respectively. $D(x)$ is the estimate of the discriminator that $x$ is real while $\mathbb{P}_r$ denotes the real data distribution. The generated data is denoted by $\tilde{x}$ with $\tilde{x}=G(z)$ and $z\sim p(z)$ where $p$ denotes some distribution, such as uniform or Gaussian. $\mathbb{P}_g$ is the distribution of the generated data or, put differently, the model distribution. The adversarial setting does not require explicit modelling of the probability distribution of the training data, which can be difficult or impossible for high-dimensional and complex data. Instead, GANs implicitly learn the distribution by generating synthetic samples that match the real ones. According to \citet{bengio_learning_2009}, the primary motivation for the design of GANs is that the learning process requires neither approximate inference nor approximation of a partition function gradient. If the objective function of the discriminator is convex, the procedure is guaranteed to converge and is asymptotically consistent. In practice, however, the above vanilla GAN suffers from stability problems and is difficult to train \citep{goodfellow_generative_2014}, especially when the discriminator's objective is not convex.
\citet{arjovsky_wasserstein_2017} address this problem by proposing a Wasserstein-GAN (WGAN), which shows a more stable training behaviour and correlation between the value of the objective and sample quality. An intuitive explanation for the Wasserstein distance as a loss function is its synonym \textit{Earth Mover Distance}. It can be related to the minimum cost of transporting mass to transform the distribution $q$ into the distribution $p$, where the cost is mass times transport distance. The Wasserstein metric changes the objective function (\ref{eq:vanillaGan}) to:
\begin{align}
\min_{G} \quad \max_{D\epsilon \mathcal{D}}\quad \mathop{\mathbb{E}}_{x\sim \mathop{\mathbb{P}_r}}[D(x)]-\mathop{\mathbb{E}}_{\tilde{x}\sim \mathop{\mathbb{P}_g}}[D(\tilde{x})]
\end{align}
where $\mathcal{D}$ is defined as a set of 1-Lipschitz functions and $\hat{x}$ is a randomly sampled subset. Instead of classifying generated images as real or fake, the WGAN-GP replaces the discriminator model with a critic that scores the realness or fakeness of an image. \citet{gulrajani_improved_2017} further improve training in their WGAN-GP by incorporating gradient penalty terms for the discriminative network. The penalty further alters the objective function to:
\begin{align}
\min_{G} \quad \max_{D\epsilon \mathcal{D}}\quad \mathop{\mathbb{E}}_{x\sim \mathop{\mathbb{P}_r}}[D(x)]-\mathop{\mathbb{E}}_{\tilde{x}\sim \mathop{\mathbb{P}_g}}[D(\tilde{x})]\\
+
\lambda \mathop{\mathbb{E}}_{\tilde{x}\sim \mathop{\mathbb{P}}_{\hat{x}}} [({\| \nabla_{\hat{x}}D(\hat{x})\|}_2-1)^2] \nonumber
\end{align}
The gradient penalty term $[({\| \nabla_{\hat{x}}D(\hat{x})\|}_2-1)^2]$ opposes vanishing and exploding gradients. The major advantage of WGAN-GP is its improved convergence. Previous models optimizing equation (2) suffer from mode collapse, a scenario in which the generative model samples with little to no diversity. Advanced methods adapt the model architecture as done by \citet{allahyani_divgan_2023} or improve the loss functions as in WGAN-GP. We refer to \citet{gulrajani_improved_2017} for more detailed explanations.\\
GANs need large amounts of training data to learn the underlying distribution of the data accurately. Insufficient or unrepresentative data produces unrealistic and low-quality synthetic samples. Further, GANs can suffer from instability and sensitivity to hyperparameter settings, making them challenging to tune and optimize. At the same time, the choice of architecture, optimization method, and regularization techniques can significantly affect their performance and stability. 

Whereas WGAN-GP relies on image input, other models rely on time series as inputs \citep{yoon2019,esteban_real-valued_2017}. These rely on recurrent neural networks (RNNs) and have already been applied in economic and financial settings for identifying trading strategies and enhancing predictive tasks in the market environments \cite{carvajal-patino_synthetic_2022,wu_prediction_2022}. 
The most prominent model is TimeGAN \cite{yoon2019}, which outperforms many related models, e.g. the ones proposed by \cite{esteban_real-valued_2017} and \cite{donahue_adversarial_2019}.
TimeGAN relies on various networks and combines supervised and unsupervised training to account for temporal correlations. TimeGAN learns a time series embedding and optimizes multiple loss functions. Its architecture comprises an encoder/decoder structure to facilitate the process and a supervising network to preserve intertemporal dynamics. Thus, TimeGAN requires five different loss functions to control the generator, the discriminator, intertemporal relationships, moments, and the embedding loss. Encoder/Decoder networks become obsolete for the WGAN-GP since the proposed image representation captures temporal dynamics and omits an embedding loss due to its numeric invertibility. 
Further, the TimeGAN architecture assigns each network a different task, and each component is trained separately before starting joint training. Despite the slow training, the TimeGAN is considered a proper benchmark for any generative time series model due to its superior simulation results \citep{yoon2019,michalowski_medical_2020,naveed2022assessing}.
\subsection{Time Series Image Representations}
Identifying an efficient representation is essential to leverage the effects of image-based GANs for time series synthesis. Transferring time series to images to infer statistical properties is a well-established and active research subject regarding generation and forecasting. All image representations for univariate time series construct a 2D encoding by performing a function $f$ on the pairwise combination of time series $x$ at two steps in time $i$ and $j$. A time series of length $S$ is transformed into a 2D matrix by 
\begin{align}
    R_{i,j}=f(x_i,x_j) \quad & \forall i,j\in \{1,2,\ldots ,S\}
\end{align}
while $f$ can perform any operations on the two inputs. However, not all operations performed on the two inputs are suitable. The choice of $f$ significantly impacts the representation's applicability and must be chosen carefully. Two criteria can classify the functions applied to the time series return representations. The first criterion is invertibility. Invertible representations allow encoding a time series in an image and retrieving the time series without any additional information. Non-invertible image representations often rely on a binary format, but even encodings in a continuous format may be non-invertible. Non-invertibility results from the function applied to the input pair $x_i$ and $x_j$  or from a missing scale required for inversion, analogue to retransforming differenced values or a series of returns to their initial values. The diagonal of an image representation determines the invertibility. For the exemplary case of summation $f(x_i,x_j)=x_i+x_j$, the diagonal is defined by $R_{i,j}=x_i+x_j=2x_i=2x_j$. With this information, we can invert the representation, as opposed to the case of a differencing approach $R_{i,j}=x_i-x_j=0$. 

The second criterion is scale-invariance, which means that an image representation captures the characteristics of a time series regardless of. A time series on different scales but with the same properties results in the same image. Note that both criteria are not mutually exclusive and can also be partially fulfilled. For the context of generating time series, a representation that encodes data in continuous values and allows for inversion while capturing properties in a scale-invariant way is a good candidate. Combining both criteria, they learn patterns across scales, allowing for easy inversion back to time series.

\section{Related Work}
\label{sec:related_work}
Three streams of literature are related to our work: research on image representations for time series, generative models, and methods for data augmentation. Below, we describe how each of the streams is affected by our contribution and discuss related approaches.
\subsection{Time-Series Image Representation}
We contribute to the current research on time series image representations in three ways. \newline
First, we propose a continuous, deterministic image representation. Previous representations, such as the recurrence plot developed in the 1980s by \citet{eckmann_recurrence_1987}, yield a binary encoding. They have found frequent applications, especially in chaos theory \citep{marwan_recurrence_2007} and system engineering \citep{SHAHVERDY2020113240} to identify the state of an underlying system. Binary encodings do not support regression but can provide additional features in forecasting tasks \citep{LI2020113680}. However, reliance on $\epsilon$, which defines a threshold value to indicate whether a point in time is a recurrence, is a drawback of binary encodings. The presence of non-stationarity makes choosing a fixed $\epsilon$ almost impossible. Our proposed method refrains from making an assumption on $\epsilon$ and instead relies on the logarithmic ratio of two points in time. Another class of representations is Markov transition fields \citep{wang2015imaging}. They bin a time series into a distinct number classes, calculate transition probabilities, and encode them in a two-dimensional space. The number of bins defines the dimension of the encoded image. Spectrograms also rely on probabilistic encoding. These are used similarly to a waveform, which processes audio signals and visualizes the spectrum of frequencies of a signal as it varies with time \citep{wyse_audio_2017}. Each plot is then distinct for a certain sound or spoken word and can be used to classify audio signals.\\
\\
Second, we devise an image representation that treats the image's diagonal differently from the off-diagonal to ensure invertibility. The only comparable invertible representations are the Gramian Angular Summation Field (GASF) \cite{wang2015imaging} and Fourier Transform Methods. Even though Fourier Transform Methods such as the Short-term Fourier Transform used by \cite{zhang_solargan_2023} in an exemplary application have proven successful, they are highly dependent on the choice of hyperparameters, such as the window size used for the transformation. This hinders immediate benchmarking due to the large parameter space. Similar to the proposed XIRP, the GASF is also a data-transformation method to convert one-dimensional time-series data into a sequential set of images that retains the temporal features of the original data while adding temporal correlations. By representing each point in time by its polar coordinates and computing a cumulative sum, it treats off-diagonal values, which capture the intertemporal dynamics the same as diagonal values. In the  empirical part of the paper, we examine whether the separate treatment of diagonal and off-diagonal elements in the proposed XIRP offers an advantage over the GASF approach. \\
\\
Third, the XIRP is more versatile than competing techniques. Binary recurrence plots, Markov Transition fields, and spectrograms are non-invertible. Therefore, they support classification problems but are inadequate for point forecasts with continuous values. Unthresholded recurrence plots and GASFs support both tasks and have been applied to regression and classification \citep{barra_deep_2020,ZHU2022115992,zhang_novel_2020}. However, disregarding scale, they can only generate differenced time series and require a start value, which restricts their application in practice. Due to the scale-invariant temporal dynamics and scale-dependent diagonal, our representation is generally applicable for regression and classification. 
 As part of the motivation for developing the XIRP, some representations are discussed more extensively in the upcoming chapter. In addition,  Table \ref{tab:characteristics_table} includes a summary of the prevalent methods and their characteristics \ref{tab:characteristics_table}.
 
\subsection{Generative Models}
In the context of generative models, the paper offers two contributions.\newline

First, embedding time series as images facilitates the use of state-of-the-art generative image models like GANs for time series synthesis, which simplifies the sampling process, reduces complexity, and raises sampling quality as we show in the empirical part of the paper. 
In the past, generative models mainly relied on RNNs. Recently, transformer models have been introduced that act adversarially as generator and discriminator models \citep{michalowski_tts-gan_2022}. With the advanced attention mechanism and design towards sequential data, the transformer architecture has already been successful in biological sequence analysis \citep{nambiar_transforming_2020} and natural language processing \citep{gillioz_overview_2020}. The results of \citep{michalowski_tts-gan_2022} indicate competitiveness to a TimeGAN but solely rely on medical data and require a more thorough empirical validation. The superior performance of TimeGAN compared to its competitors WaveNet \citep{oord2016wavenet} and RC-GAN \citep{esteban_real-valued_2017} results from additional networks that perform auto-encoding steps on a latent space as well as forecasting the latent space over time. The benefit of superior simulation results comes at a high cost in computing time, additional complexity, and hyperparameters.
Generative models for time series relying on image representations have been proposed by \cite{donahue_adversarial_2019}, who transferred raw audio data to spectrograms and then used a generative model based on convolutional neural networks.
For the application of time series synthesis, however, existing CNN-based generative models are outperformed by RNN-based alternatives and still suffer from the approximate invertibility of their representation. With the  proposed here, there is no need for approximation, additional auto-encoding, or latent space supervision steps.\\

Second, we enable a large class of other generative models to benefit from the advantages of the XIRP. 
For example, apart from GANs, diffusion models from thermodynamics have been successfully applied to synthesize images \citep{DURALL2023105377}. Instead of relying on an adversarial loss, their concept sequentially adds and removes noise in an input image. The sampling process is then analog to the GAN framework, moving from a noisy image toward a synthetic instance. Since diffusion models are a relatively new concept, especially for time series settings, the research is still in an early stage.
Another methodologically related stream of research examines the imputation of missing data in a probabilistic fashion \citep{alcaraz_diffusion-based_2022,park2022neural,tashiro2021}. These studies emphasize local coherence conditioned on an input series, which differs from our focus on unconditionally generating new time series.

A further merit of our approach may be seen in the fact that it could pave the way for generative models for multivariate time series synthesis, a field that is still in its infancy. In theory, our image-based approach generalizes to the multivariate case. For example, WGAN-GP accepts stacked representations as joint input for the discriminator. We suspect, however, that in high-dimensional settings, the computational performance of the WGAN-GP is affected since every time series would require a new channel. Even though we do not consider the multivariate setting in this paper, our approach might provide good results for lower dimensions. However, this still has to be proven empirically. 

\subsection{Data Augmentation}
The last stream of literature influenced by our paper is data augmentation. Our model complements existing augmentation methods with an adversarial learning method. Since the GAN samples from a noisy input, it opposes many techniques that transform the time domain by directly modifying the input time series, adding Gaussian noise or more complicated noise patterns such as spike, step-like trend, and slope-like trend \citep{Wen_2021}. Since it is non-parametric, it differs from the more advanced augmentation approaches that rely on decomposition and statistical generative model learning methods. Even though we also model a conditional distribution, our approach stands out from other conditional methods that typically involve modelling the dynamics of the time series with statistical models, e.g. Gaussian trees \citep{gaussiantrees2014} by conditioning it on noise instead of the previous point in time. Overall, the GAN-based approach complements the augmentation methods by a new category instead of being assignable to an existing one.\\
At this point, image-based generative processes often require an additional network to approximate the inversion of the representation. An invertible representation can simplify and improve the generative process via an adversarial approach with only two networks. An off-the-shelf WGAN-GP allows for quick training and may scale to higher dimensions. In sum, the WGAN-GP approach extends the set of data augmentation methods with the chance of enriching the training data and supports the model's generalization capabilities.
\section{Methodology}
This section briefly reviews existing image representations for time series before developing the ideas of IRP and XIRP.\newline
\citet{eckmann_recurrence_1987} introduced the first time series image representation. Their recurrence plot depicts points where the time series has approximately the same value. The representation is widely used and obtained as follows:
\begin{equation}
R^{B}_{i,j}={\begin{cases}1 \quad {\text{if}}\quad \|x_i-x_j\|\leq \varepsilon \\0 \quad {\text{otherwise}},\end{cases}}
\quad \forall i,j\in \{1,2,\ldots ,S\}
\end{equation}
with $R^{B}_{i,j}$ being an $S$x$S$ matrix with indexes $i$ and $j$. The representation is similar to a (binary) correlation matrix where each step in time corresponds to a feature, and $S$ corresponds to the sequence length. A recurrence occurs if the absolute difference between the values at time $i$ and $j$ is smaller than a threshold $\epsilon$. By definition, the diagonal $d$ of $R^{B}_{i,j}$ is filled with recurrences ${\forall i,j\in \{1,2,\ldots ,S\},i = j\implies d_{i,j}=1}$. Consideration of the absolute difference also implies symmetry, $R^{B}_{i,j}= R^{B}_{j,i}$. \newline 
Recurrence plots incorporate significant statistical properties, including  trends, jumps, and cyclic behaviour. To extract such characteristics, recurrence plot quantification analysis (RQA) offers several proxies for the predictability of the underlying process or how long it remains in a particular state \citep{marwan_recurrence_2007}. \newline
Recurrence plots are powerful for time series analysis but require adjustments to serve as image representation for sampling synthetic time series. Encoding whether an absolute difference of times series values at index $i$ and $j$ exceeds $\epsilon$, "non-recurrences" can hypothetically take any value. This implies that binary recurrence plots are not invertible and cannot serve as input to a time series generation process.  
\newline
We compute the difference between values at different time points or their ratio to ensure that a time series image representation is reversible. 
Differencing a time series stabilizes the mean and mitigates the impact of trend and seasonality \citep{sutcliffe_time-series_1994}. Generally speaking, differencing transforms a time series towards stationarity. This concept is popular in economics and quantitative finance and works well in many applications \citep{TAYLOR20191193,TRAN2022102574}. To allow for a more granular approach than binary encoding, an unthresholded recurrence plot (URP), sometimes termed distance plot, can be obtained as follows \cite{iwanski1998recurrence}: 
\begin{align}
R^{U}_{i,j} = \| x_i - x_j \| \quad \forall i,j\in \{1,2,\ldots ,S\}
\end{align}
The URP conveys an idea of the magnitude by which $x_i$ and $x_j$ differ and thus adds additional information to the representation. Even though it improves upon $R^{B}$, the approach has three drawbacks. First, the differencing approach is acceptable if the feature scales do not change drastically. In cases where the values are naturally in persistent bounds, differencing is legitimate and likely to return good results if the time series evolves. However, if the time series evolves and the mean of the feature jumps to a higher level or exceeds its previous bounds, the relations discovered for past differences on an absolute level will no longer hold. Therefore, the representation $R^{U}_{i,j}$ is not scale-invariant, which decreases the chance of obtaining a robust representation across scales. Second, the difference plot does not contain the scale of the feature, which vanishes by subtracting. A missing scale makes it impossible to put the changes into perspective, leading to the third drawback. The absolute values make it impossible to use the representation in a generative setting since the initial sign of the change is unknown.
In case one would calculate an alternative representation of $\hat{R^{U}}_{i,j} = \| x_i - x_j \|$, the diagonal contains $\hat{R^{U}}_{i,i} = 2x_i$ and would be invertible but still not scale invariant. 

\begin{figure*}[htbp]
    \begin{minipage}{0.32\linewidth}
        \centering
        \includegraphics[scale=.2]{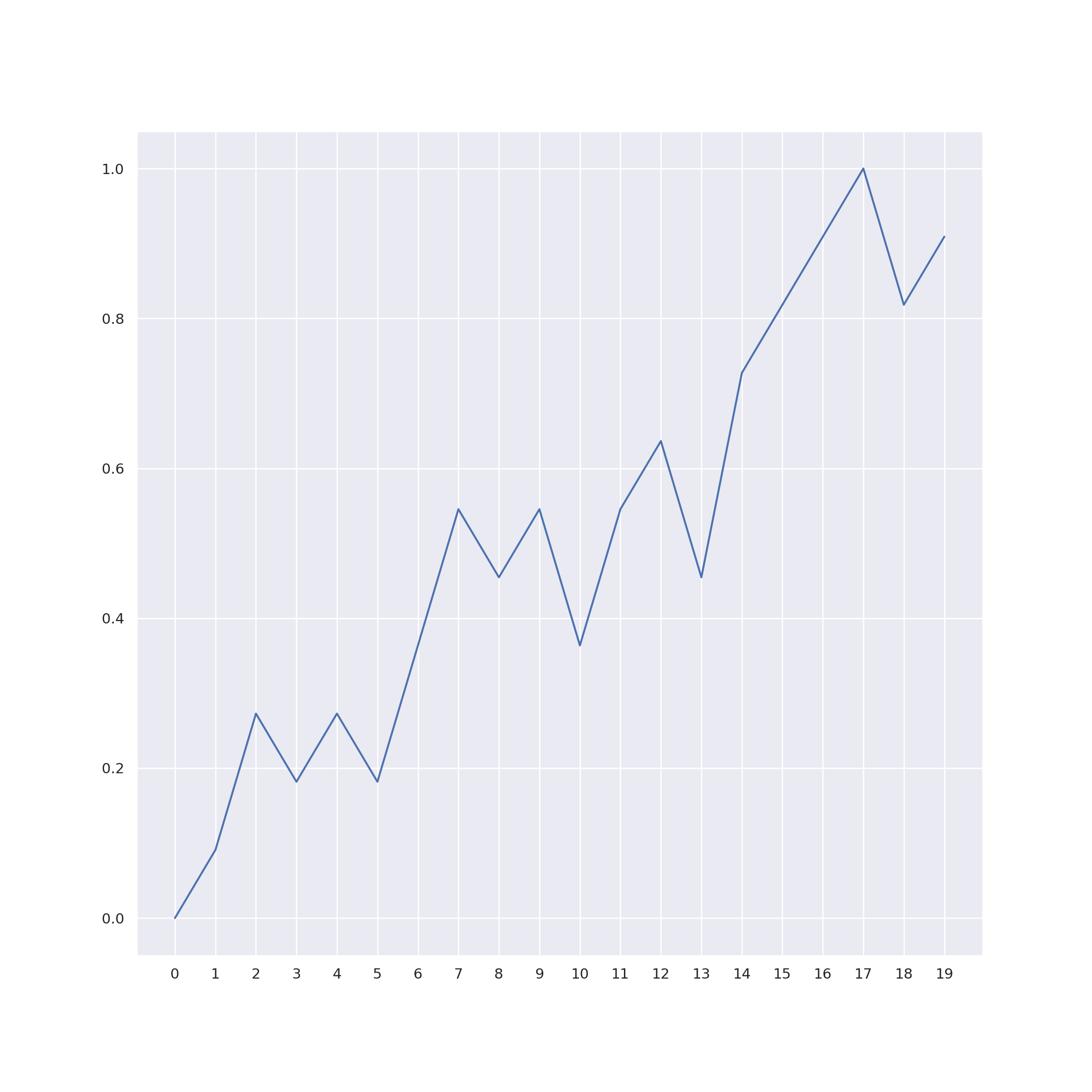}\\
         (a) AR(1) Time Series
    \end{minipage}
    \begin{minipage}{0.32\linewidth}
        \centering
        \includegraphics[scale=.2]{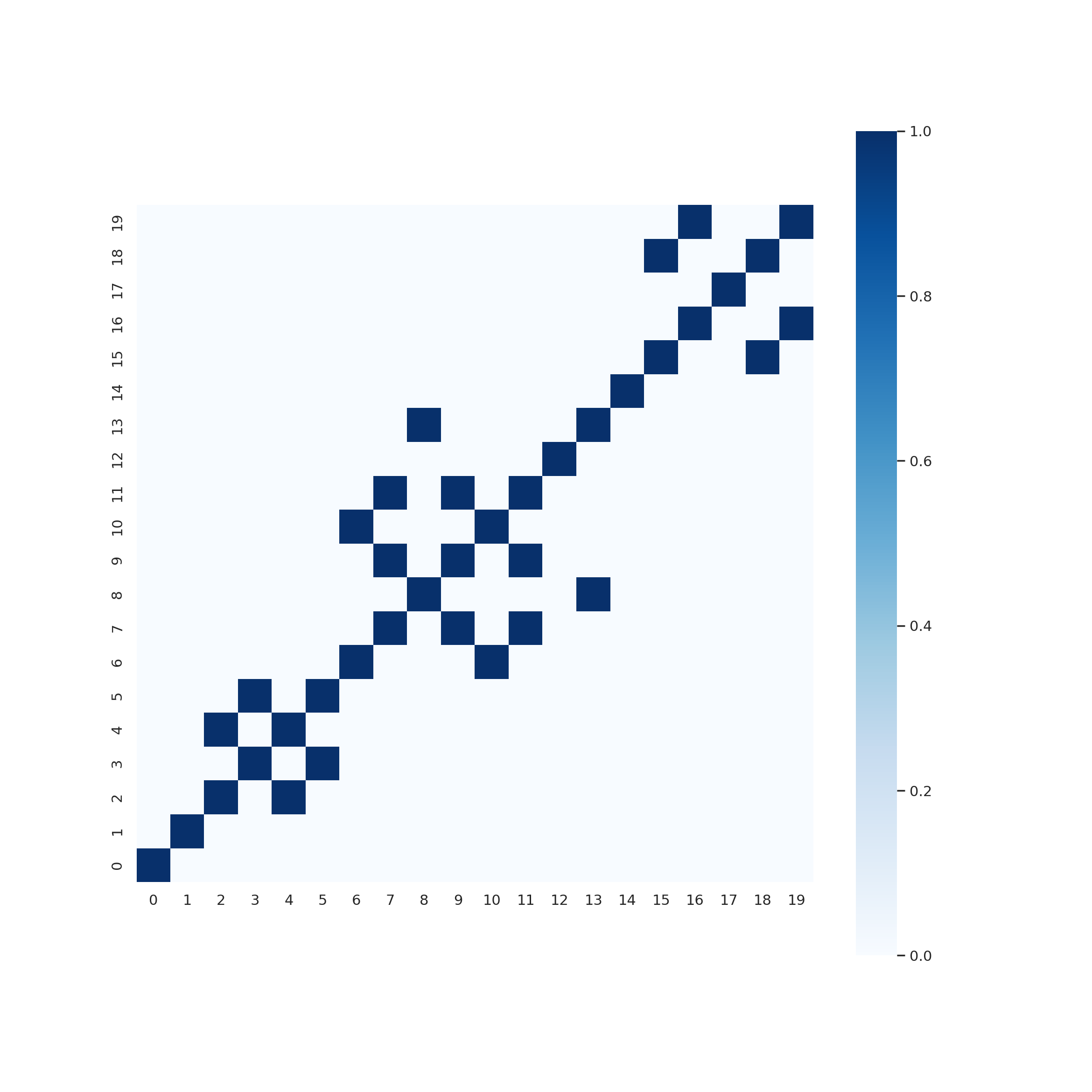}\\
        (b) Recurrence plot with $\epsilon=.2$
    \end{minipage}
    \begin{minipage}{0.32\linewidth}
        \centering
        \includegraphics[scale=.2]{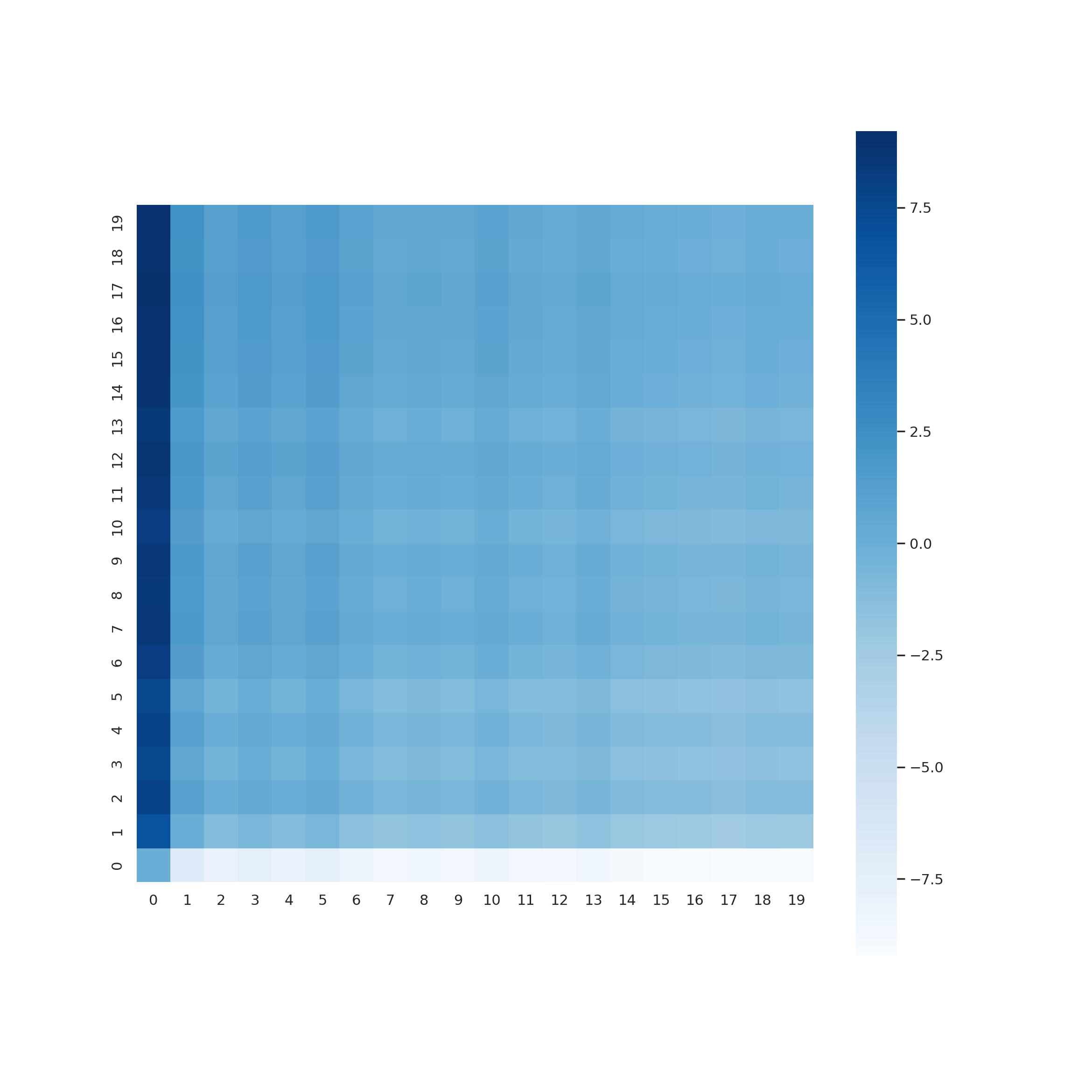}\\
        (c) Intertemporal Return Plot (IRP)
    \end{minipage}
    \\
    \begin{minipage}{0.32\linewidth}
        \centering
        \includegraphics[scale=.2]{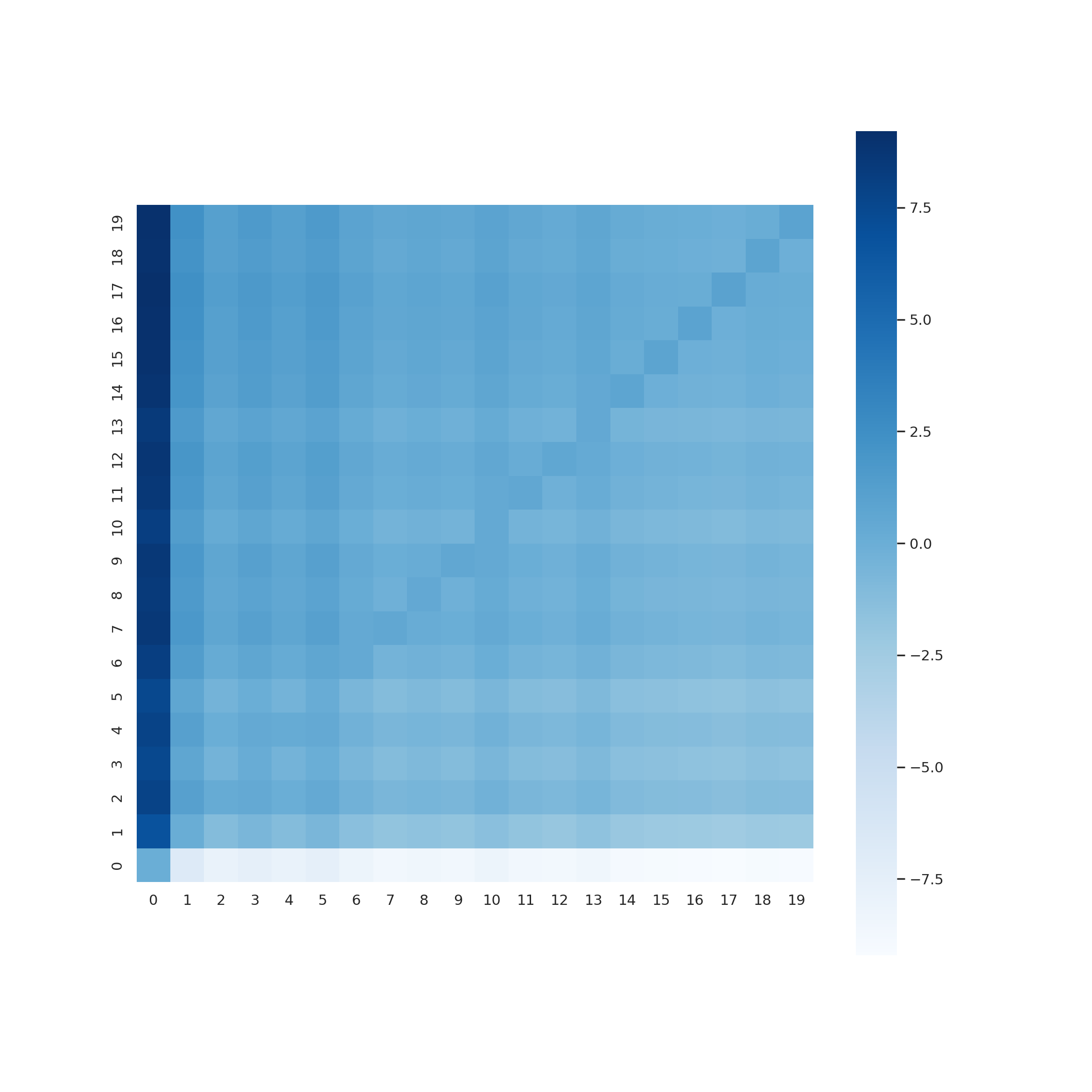}\\
        (d) Extended IRP (XIRP)
    \end{minipage}
    \begin{minipage}{0.32\linewidth}
        \centering
        \includegraphics[scale=.2,]{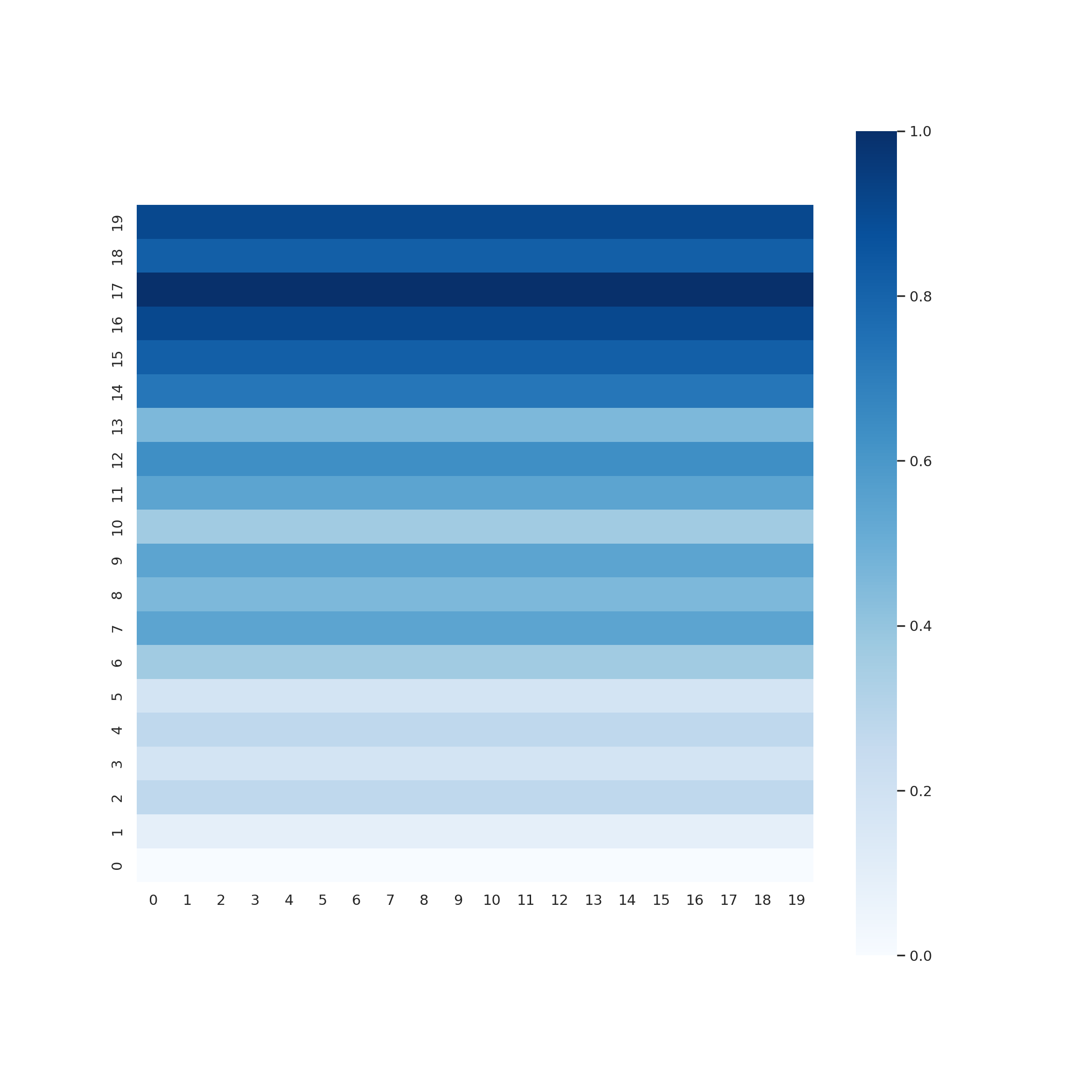}\\
        (e) Naive Representation Plot 
    \end{minipage}
    \begin{minipage}{0.32\linewidth}
        \centering
        \includegraphics[scale=.2]{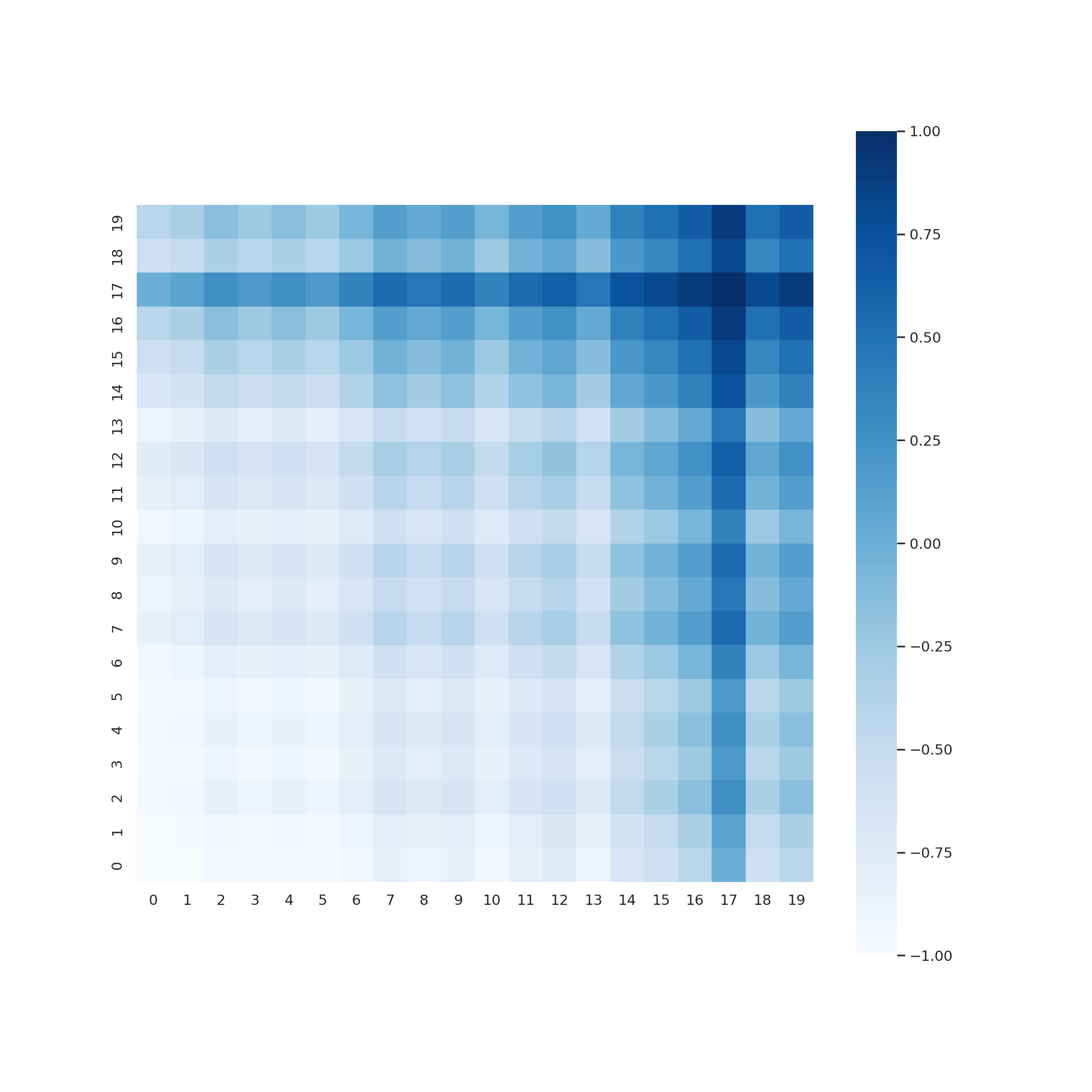}\\
        (f) Gramian Angular Summation Field (GASF)
    \end{minipage}
    \caption{AR time series with binary and intertemporal return plot}
    \label{fig:AR_recurrence_return}    
\end{figure*}

The alternative approach of using (logarithmic) ratios is frequently used in financial applications. The log transformation has a similar effect on the time series as differencing. However, calculating returns has the advantage of being invariant to scale and is especially useful when the time series evolves. Further, log returns have the desirable statistical characteristics of symmetry and additivity \citep[p.~295]{wilmott_paul_2006}.
Motivated by the benefits of a scale-invariant representation, we propose the \textit{intertemporal return plot} $R^{I}_{i,j}$ defined by 
\begin{align}
R^{I}_{i,j} =
log \left(\frac{x_i}{x_j}\right) \quad \forall i,j\in \{1,2,\ldots ,S\}
\end{align}
An advantage of $R^{I}$ over $R^{B}$ is the absence of $\epsilon$, the correct specification of which is complicated and may change over time. Further, $R^{I}$ indicates the direction compensating the limitations of $R^{U}$ while remaining scale invariant. However, the $R^{I}$ also allows inverting the time series without a specific start value. A careful look at $R^{I}$ reveals that in case $i=j \implies R^{I}_{i,j} = 0$, which leads to a diagonal filled with zeros. This inefficiency motivates the formulation of the extended IRP (XIRP), which complements the intertemporal return structure by adding extra information to the encoding. Thus, equation (3) evolves to 
\begin{equation}
R^{X}_{i,j} =
\begin{cases}
log \left(\frac{x_i}{x_j}\right) \quad & \forall i,j\in \{1,2,\ldots ,S\} \wedge i \neq j \\
x_{i} \quad & \forall i,j\in \{1,2,\ldots ,S\} \wedge i=j
\end{cases}
\end{equation} 
The previously unused diagonal of the IRP fills with additional information. $R^{X}$ now has all relevant characteristics regarding scale-invariance and invertibility required for a generative model.
Another approach that provides both scale-invariance and invertibility is the Gramian Angular Summation Field (GASF) concept. It combines the previously mentioned benefit of summation proposed for $R^{U}$ with bounded feature scales. The GASF representation has already been tested extensively in various applications, from forecasting sun flares \citep{chen_effectiveness_2020, Nagem2018PredictingSF} to stock prices
\citep{LIANG2022117595}. It transfers a [0,1] scaled time series $x$ for a point $t$ in time to 
\begin{align}
\phi_t = arccos(x_t)
\end{align}
which allows calculating the GASF for two points in time via
\begin{align}
R^{GASF}_{i,j} = cos(\phi_i+\phi_j) \quad & \forall i,j\in \{1,2,\ldots ,S\}
\end{align}
According to \cite{wang2015imaging}, it provides a way to preserve the temporal dependency since time increases as the position moves from top-left to bottom-right in $R^{GASF}_{i,j}$ and the main diagonal allows to reconstruct the time series. Even though it scales the input beforehand, it is not scale-invariant since multiplying the scaled input by a constant returns a different GASF in the off-diagonals. This is considered a drawback since the dynamics remain unchanged but are encoded differently in the GASF.
A last representation introduced mainly for benchmarking purposes is the naive representation defined by
\begin{align}
R^{N}_{i,j} = x_i \quad & \forall i\in \{1,2,\ldots ,S\}
\end{align}
Technically, the representation concatenates the time series columns $S$ times, returning an image where each row has the same value. The naive representation removes all additional information that could originate from an advanced representation. At this point, there are two reasons why we need a naive baseline representation to compare GASF and XIRP to the results of the TimeGAN. First, we would like to identify the benefit of each representation for measures of similarity and predictive ability compared to the naive representation. Second, we are comparing two inherently different models in the benchmark section. The TimeGAN has five loss functions compared to a WGAN-GP with one loss function. We use the naive representation to reduce the possible benefit of a superior representation, so neither TimeGAN nor WGAN-GP has proprietary information.
Figure \ref{fig:AR_recurrence_return} (a) displays a random time series of $S=20$ steps with a positive trend. For the recurrence plot in Figure (b), the dots mark the recurrence and cluster around the diagonal, which is zero by definition. Clusters around the diagonal indicate a drift in the time series since recurrences only appear in short periods since the time series evolves. The IRP and XIRP look similar with the difference of the diagonal, which is 0 in the IRP and contains the initial time series values from Figure \ref{fig:AR_recurrence_return} (a) in the diagonal. Both plots show a significant return at the beginning of the time series and smaller ones in the later course. The naive representation in Figure \ref{fig:AR_recurrence_return} (e) shows stripes resulting from the time series' column-wise concatenation. The GASF displayed in Figure \ref{fig:AR_recurrence_return} (f) shows a contrasting pattern assigning smaller values to the lower left corner and higher values to the upper right since it refers to the polar encodings and not its returns.
The naive representation completes the set of candidates for the subsequent benchmark. Linking the concepts of GAN and XIRP/GASF, we can sample and extract the time series from the diagonal and the temporal dependencies in the off-diagonal.\newline
There are multiple ways to invert the underlying time series from both representations. Since the diagonal contains the initial values, they are extractable without considering the off-diagonal. Although this inversion is most accurate numerically, it would neglect parts of the encoded information and lead to inferior simulation results. Thus, we propose to extract the diagonal in the first step and then use the diagonal as a baseline time series to calculate variants via the off-diagonal returns. Once all $n$ series are inverted, there exists the possibility to either average all series, which we refer to as \textit{inversion by mean} (IM), or to randomly select an instance, which we label \textit{inversion by random column} (IRC). Note that both IM and IRC are a form of approximate inversions.
A simple example can illustrate the approach with the help of Figure 1. In the case of the Naive representation, the IM method averages over each row of the plot, obtaining one value for each line. In the case of IRC, a column is selected randomly.\newline
\section{Experiments}
\label{sec:experiments}
In the following, we first revisit our research questions and describe how each experiment contributes answers. We next outline how qualitative and quantitative experiments cover different aspects of time series synthesis quality and introduce an evaluation scheme for the obtained results. Then, we briefly present the synthetic and real data sets, outline their properties and motivate our choice to incorporate them in the train data.

\subsection{Design}
The key questions to answer by empirical experimentation are 1) which image representation is better suited to capture temporal dynamics and 2) whether the proposed pipeline of time series image representation, image generation, and image inversion is competitive to a state-of-the-art generative time series model? \newline 
To answer both questions, we conduct a series of experiments starting with a qualitative visual inspection of simulation results. 
For the visual inspection, we first calculate image representations for a set of real-time series. Next, we train WGAN-GP to simulate synthetic XIRPs, GASFs, and naive representations. Finally, we apply a clustering algorithm to facilitate visual comparisons of synthetic and original time series in a two-dimensional space. A single dot represents an original or synthetic time series, depending on its color. Elements located closely in the two-dimensional space will likely have similar characteristics in the higher-dimensional space. The method used for clustering is the Uniform Manifold Approximation and Projection (UMAP) algorithm \citep{mcinnes_umap_2020}, which interprets the $d$ time steps as an individual feature. \cite{yoon2019,ali_timecluster_2019} use a similar approach for clustering time series. In case original and synthetic elements differ, the UMAP algorithm forms separate, homogeneously coloured clusters with little distance between the elements inside each cluster and large distances between the individual clusters. Colouring the elements in the plot visualizes their class membership of either original (brown) or synthetic (blue). The algorithm returns heterogeneous clusters of mixed red and green dots if the elements are similar.
For the next set of experiments, we require quantifiable and comparable measures of model performance. The goal is to measure the similarity and predictive ability of each model, representation, and inversion method. The first backtest trains a model on synthetic data to predict original data. The intuition is that if the synthetic data is close to the original data, the trained model should be able to predict the original data. Thus, it measures the predictive performance of the generative model. We train a gated recurrent unit (GRU) on synthetic data sequences from $t=0$ up to $t=d-1$ and predict each sequence's $d^{th}$ element. We repeat the process $k=10$ times to obtain reliable results and apply early stopping as a regularization technique before calculating the MAE on a test set of original data. We then average the MAE over $k$ folds. The lower the MAE, the better the model synthesizes the data. \newline
In a second discriminative test, we use each generative model to generate synthetic data and let a GRU discriminate between original and synthetic data. In this test, lower (higher) classification accuracy indicates high (low) similarity between the synthetic and original data. Thus, the test aims at judging the similarity between synthetic and original elements. A higher binary classification error signals a higher quality of generated data. Again, we use early stopping and a test set for all $k=10$ iterations before averaging the binary cross-entropy error. 
Following the two backtests, an aggregation metric counts each model's best scores. It compensates for unfortunate situations in which substantial errors on a single time series have strongly impacted the average performance. Counting every best score rules out the effect of extreme scores. Further, we use a ranking method to compare the models and representations. Since not only the best model is of interest but also their relative performance, we assign rank 1 to the best and rank 4 to the worst model. Backed by the MSE of both backtests, the number of best scores and ranking, we can make a confident statement on the representations and models.\newline
Since a benchmark study across models, representations, and datasets inherits complexity, we evaluate the results concerning different aspects. We choose TimeGAN as a reference model since its results are validated and a widely used benchmark.
The TimeGAN trains 10K epochs for our benchmark test on each data set with standard configuration from the \citet{yoon2019} paper. The WGAN-GP uses standard configuration only, except for lowering the batch size from $b_{old}=16$ to $b_{new}=8$ to support generalization. The WGAN-GP and TimeGAN models train on 100 time series from 9 data sets varying in sizes from 5 to 27. Due to the univariate approach of the WGAN, the TimeGAN also trains on univariate time series.
\subsection{Data}
For our experiments, we use publicly available real-world and synthetic data. In total, 100 time series from 9 data sets test the generative capabilities of each model and representation. A limit of 1000 observations is set for each time series to speed up training and facilitate comparability between the scores among datasets. \newline
The first synthetic datasets are the sine and noisy sine datasets. They consist of 9 time series of sine curves with varying shifts and levels of noise. They are incorporated to check how well the models and representations handle (noisy) cyclic behaviour. Another critical aspect of a generative model is its handling of randomness and Markov properties, both incorporated in Brownian motion. The trajectory undergoing Brownian motion is unpredictable and exhibits erratic movements. It is a stochastic process where the future behaviour depends only on its current state and is independent of its past trajectory. Each change in the direction is independent of previous changes. To test if a model successfully captures this behaviour, we construct a Brownian motion data set with time series of varying coefficients for $\sigma$ ranging from $\sigma=.9$ to $\sigma=.99$ and $\mu=0$. Evaluating the performance of generative models under disruptive circumstances is the next point of interest. In reality, we often observe time series with jumps in the context of news releases, economic shocks, or other sudden changes in market conditions, which a Merton jump process can model. It includes a jump intensity parameter $\lambda$ that governs the frequency and magnitude of the jumps and the volatility of the diffusion process $\sigma$. To test the capabilities of models and representations concerning jump processes, the Merton Process dataset contains 10 time series with varying values for $\lambda$ and 
component $\sigma$. Another characteristic commonly observed in real-world data concerns heavy-tailed distributions, implying a slower decay than distributions with finite moments. The heavy tail suggests that rare events with large magnitudes are more likely to occur under a normal distribution than expected. We test a process where the change is governed by a power law function in which a slight increase is typical while more significant increases are rare. In the corresponding experiments, $\alpha$ is in the range of [0.1,0.9]. \newline
Next, we consider the real-world datasets starting the UCI Appliances energy prediction \citep{Dua2019} dataset, characterized by noisy periodicity. The data consists of multivariate, continuous-valued measurements, including numerous temporal features measured at close intervals. By contrast, sequences of the stock price dataset are continuous-valued but aperiodic. We use the first 1000 trading days of historical Google stocks data from 2004 to 2008, including volume and high, low, opening, closing, and adjusted closing prices. The air quality dataset contains the responses of gas sensors near an Italian city containing hourly response averages recorded along with gas concentrations \cite{Dua2019}. The dataset combines characteristics of the synthetic datasets and includes a total of 13 time series with different properties in terms of periodicity and jumps. The bike share dataset combines the seasonal patterns with irregular jumps for features such as weather conditions, precipitation, day of the week, season, hour of the day, and rentals by hour. The exact number of features $n$ for each dataset can be obtained from Table \ref{tab:result_table}.

\section{Empirical Results}
In the first part of the empirical results, we discuss three arbitrarily chosen synthesis results from WGAN-GP and TimeGAN. This is for illustration. Recall that the total number of combinations of XIRP, GASF, and naive representation with recovery methods IRC and IM accumulates to $n=600$. Hence, it is impossible to analyze all samples visually. For every UMAP plot, we randomly chose 500 original and synthetic elements.
\begin{figure*}
    \begin{subfigure}{.49\linewidth}
    \centering
    \captionsetup{skip=-10pt}
    {\includegraphics[scale=.29,trim={.4cm .4cm .4cm .4cm}]{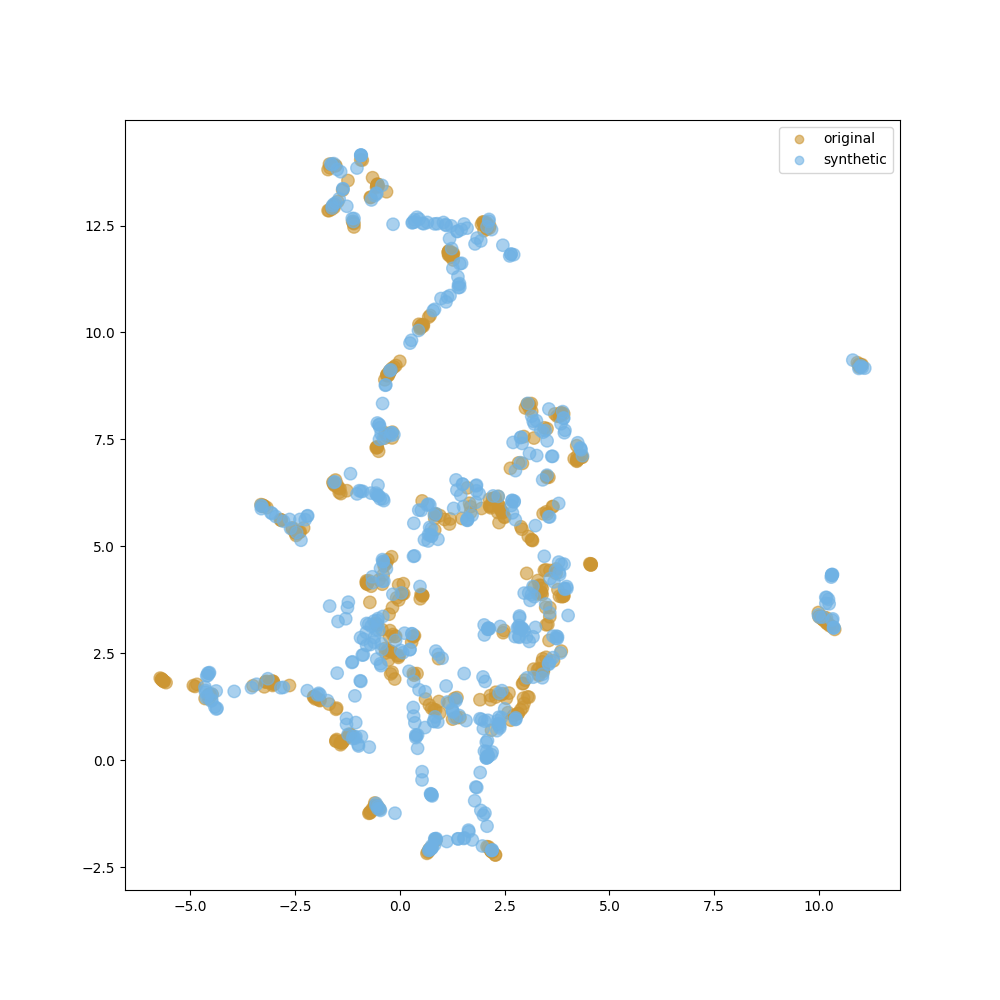}}
    \caption{WGAN-GP on bike share \textit{cnt} \\(XIRP, IRC)}
    \end{subfigure}
    \begin{subfigure}{.49\linewidth}
    \centering
    \captionsetup{skip=-10pt}
    {\includegraphics[scale=.29,trim={.4cm .4cm .4cm .4cm}]{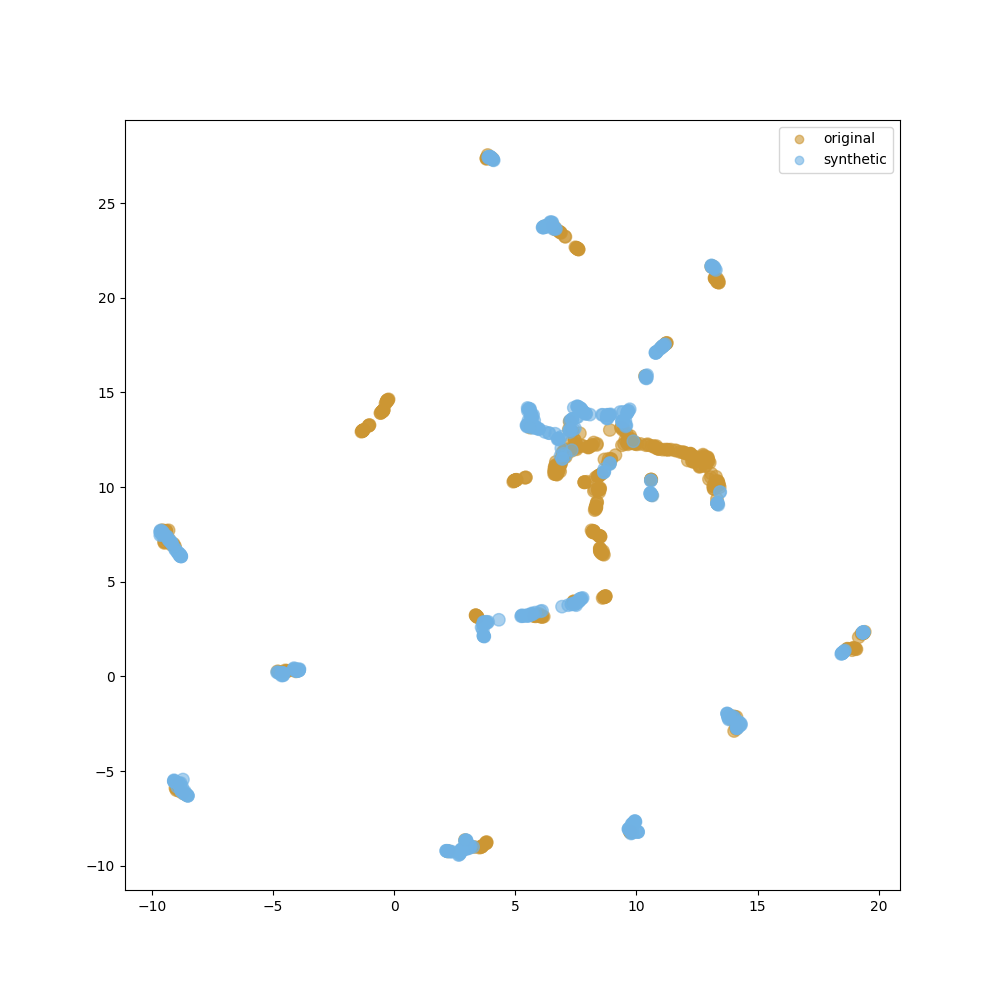}}
    \caption{TimeGAN on bike share \textit{cnt}\\ $\quad$}
    \end{subfigure}

    \begin{subfigure}{.49\linewidth}
    \centering
    \captionsetup{skip=-10pt}
    {\includegraphics[scale=.29,trim={.4cm .4cm .4cm .4cm}]{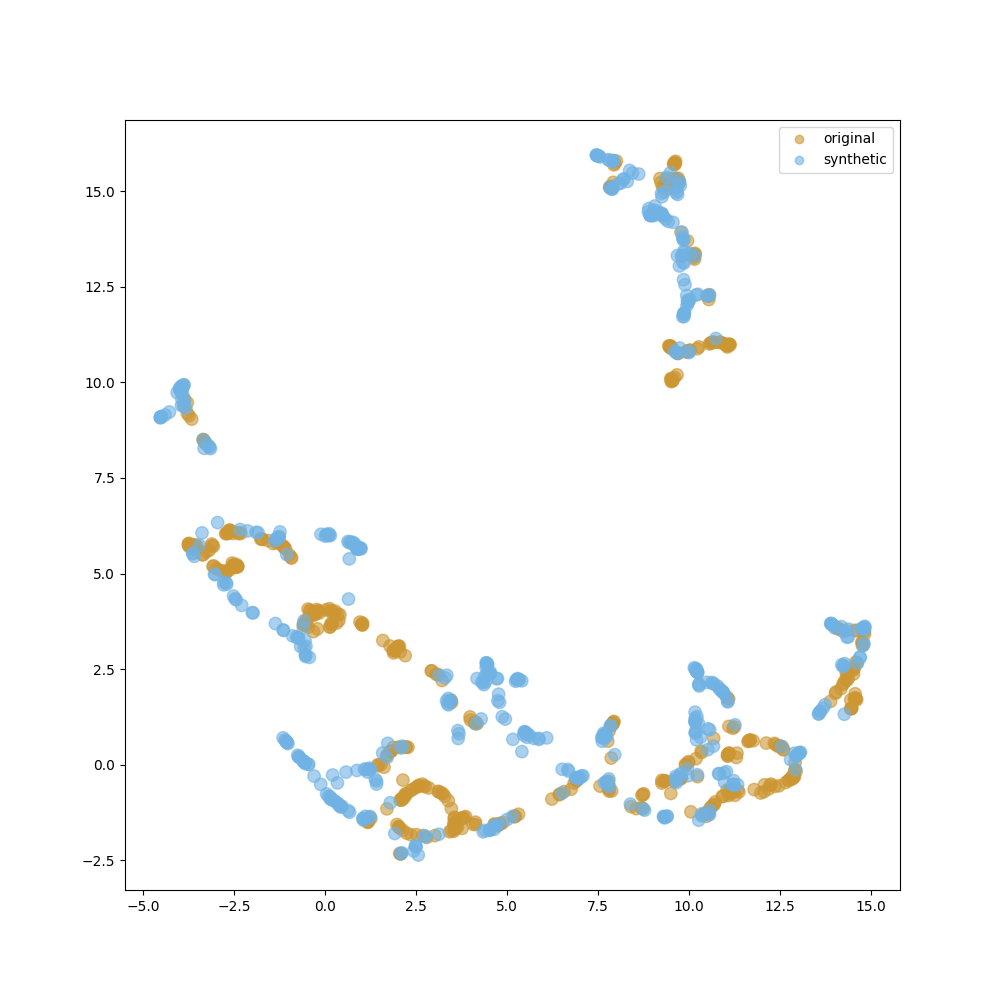}}
    \caption{WGAN-GP on stock data \\ \textit{Adjusted Close} (XIRP, IRC)}
    \end{subfigure}
    \begin{subfigure}{.49\linewidth}
    \centering
    \captionsetup{skip=-10pt}
    {\includegraphics[scale=.29,trim={.4cm .4cm .4cm .4cm}]{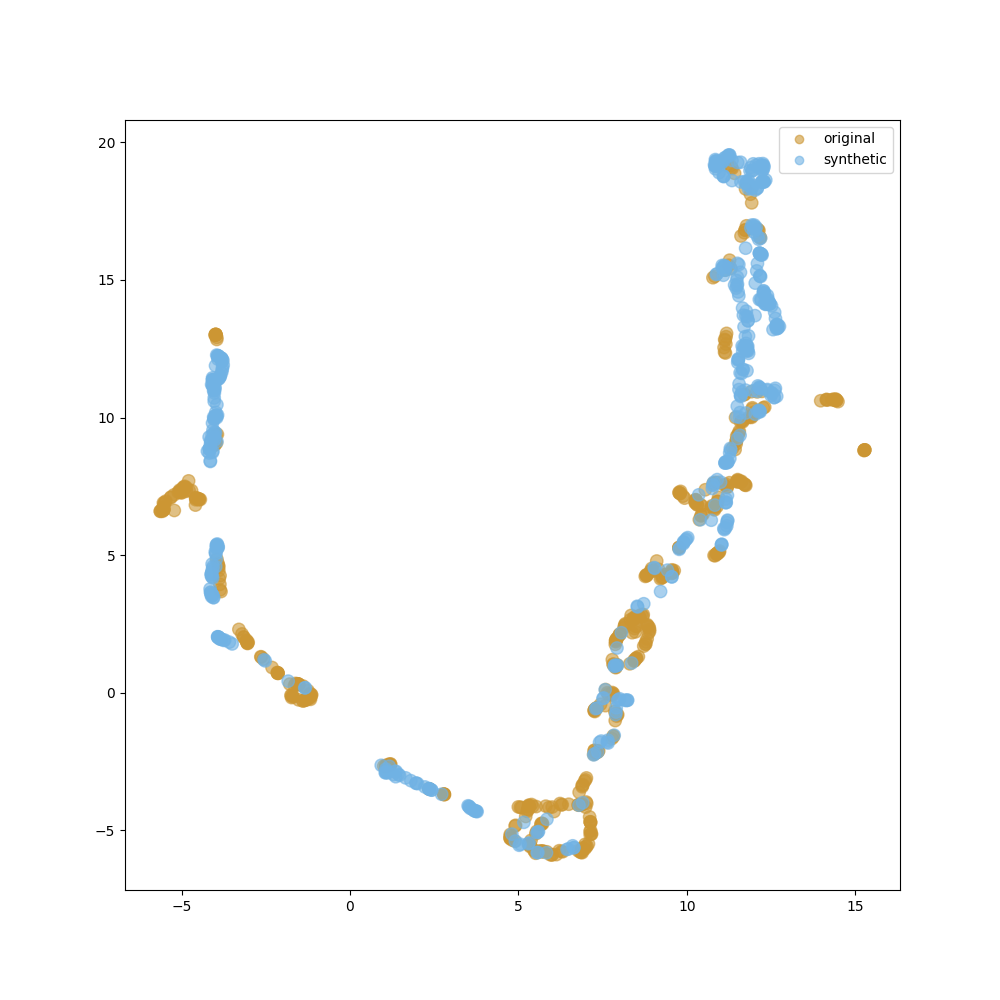}}
    \caption{TimeGAN on stock data \\\textit{Adjusted Close}}
    \end{subfigure}
    
    \begin{subfigure}{.49\linewidth}
    \centering
    \captionsetup{skip=-10pt}
    {\includegraphics[scale=.29,trim={.4cm .4cm .4cm .4cm}]{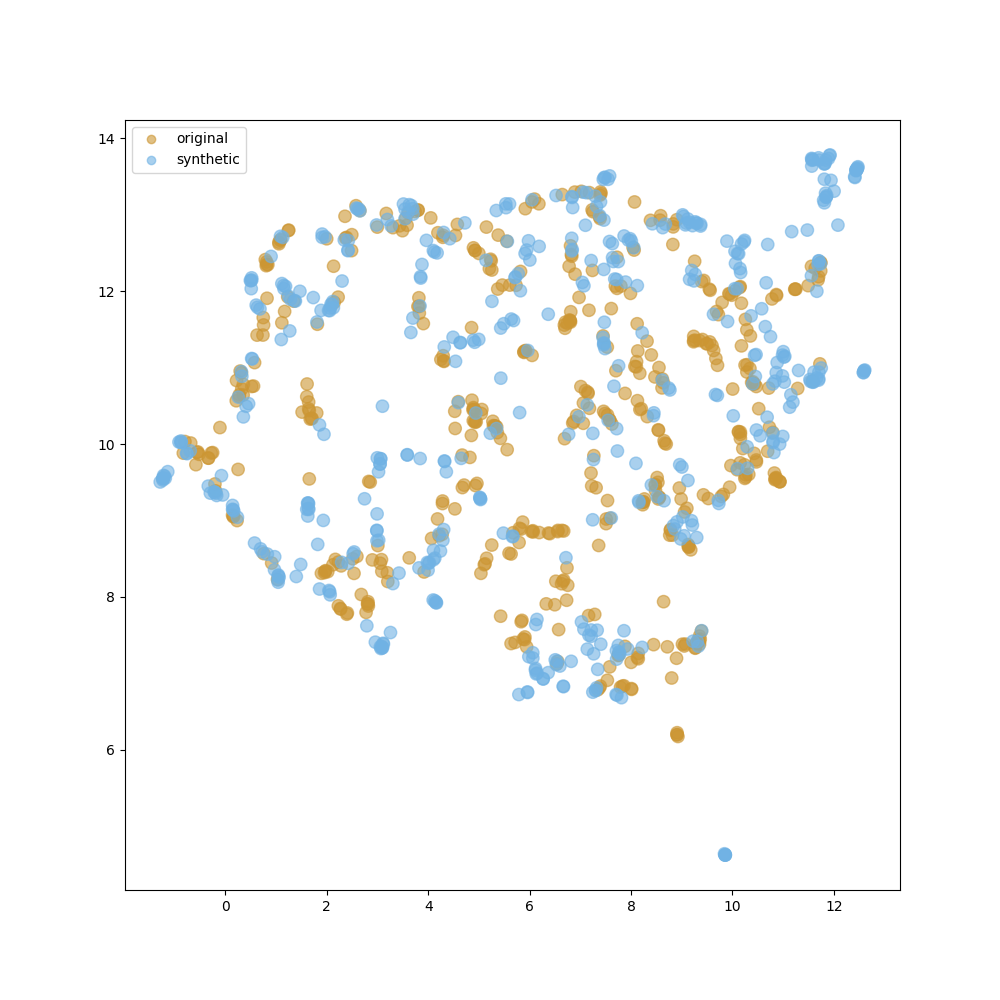}}
    \caption{WGAN-GP on \textit{hum} of bike share (XIRP, IRC)}
    \end{subfigure}
    \begin{subfigure}{.49\linewidth}
    \centering
    \captionsetup{skip=-10pt}
    {\includegraphics[scale=.29,trim={.4cm .4cm .4cm .4cm}]{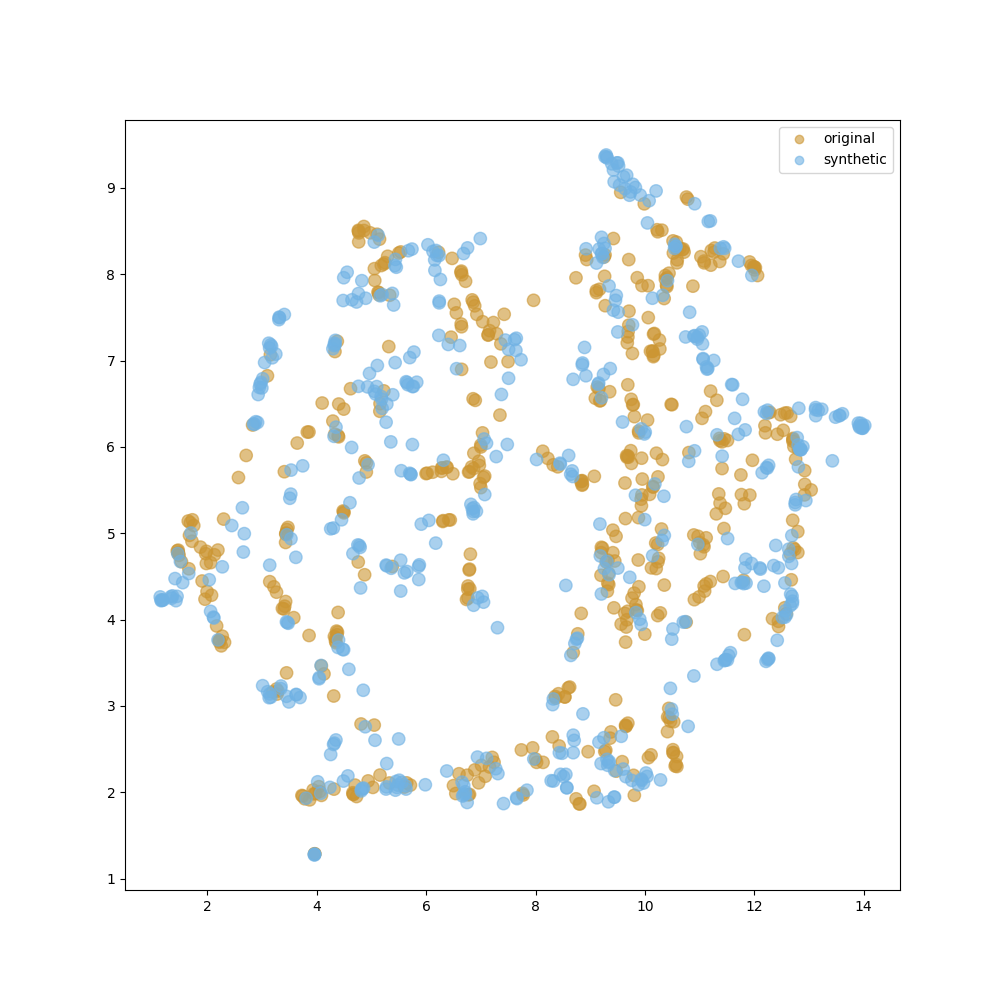}}
    \caption{WGAN-GP on \textit{hum} of bike share (GASF, IRC)}
    \end{subfigure}
    \caption{UMAP plots for real and synthetic Time Series}
    \label{fig:UMAP}%
\end{figure*}

The first example displayed in Figure \ref{fig:UMAP} (a) shows original and synthetic elements generated by WGAN-GP on the count of total rental bikes \textit{cnt} of the bike share dataset. The original elements marked in brown predominantly cluster in the left half, consisting of a circular shape in the centre and two curved shapes to the left and top. Further, a small dense cluster forms to the upper right of the circular shape. A few isolated original synthetic points spread on the right. The synthetic elements distribute well between the original elements and come close to their shape. Surprisingly, the synthesis covers even the remote real elements on the right. Synthetic elements below the circle of real instances bridge the gap between real and synthetic instances, proposing new plausible shapes. From a visual inspection, synthetic elements augment the data well.
Figure \ref{fig:UMAP} (b) displays the result for \textit{cnt} for the TimeGAN model. Original instances mainly populate the centre of the UMAP. We also observe more separate clusters mixed with synthetic and original instances. The visual inspection suggests that the TimeGAN model performs worse for this data set than the WGAN-GP. The shapes compared to (a) differ significantly due to the random choice of the instances and the difference in the synthesis result. Since the UMAP uses randomness for fast approximation and solving optimization problems, different runs of UMAP can produce different results.
Figure \ref{fig:UMAP} (c) depicts the augmentation by a WGAN-GP with XIRP on the \textit{Adjusted Close} of the stock dataset. 
From a visual inspection,  synthetic elements augment the data reasonably well, to the extent that there are no remote clusters of synthetic or original elements. Around the origin, the synthetic data expands the existing original data points. Also, the synthetic elements spread evenly among the original instances in other plot areas, indicating a good augmentation quality. Another successful augmentation can be found in Figure \ref{fig:UMAP} (d). Here, the TimeGAN algorithm fits the synthetic elements close to the original elements, forming a shape similar to beads on a string. However, some areas remain uncovered, especially on the left and right. Further, we can see that synthetic elements overpopulate the upper right corner, while the bottom of the v-shaped cluster contains only a few instances. However, from the visual inspection, it is impossible to say whether this effect is structural or random due to the sample chosen. 
Figure \ref{fig:UMAP} (e) shows a WGAN-GP with XIRP on the humidity time series \textit{hum} using IRC. Both synthetic and original elements are well mixed in a single central cluster. Single isolated areas with synthetic elements can be observed only in the upper right corner, and one synthetic element at the bottom. Otherwise, the synthesis results strongly resemble the original data. Comparing the results to another image representation, namely GASF, in Figure \ref{fig:UMAP} (f) obtains similar results. Analogously, the original and synthetic elements for the GASF representation cluster centrally but with a slightly different shape. Again, small areas with only synthetic elements are at the top and right. 
The small areas we see in Figures \ref{fig:UMAP} (e) and \ref{fig:UMAP} (f) where only original or synthetic elements cluster can result from the selection of elements and need not necessarily result from a poor synthesis.
\newline
Regarding question 1) which image representation is better suited to capture temporal dynamics, we cannot make a definitive statement at this  stage. For the XIRP and GASF, we observe successful (and imperfect) synthesis results. Both representations catch the dynamics well and deliver good results according to the UMAP algorithm. To answer the second research question, whether an image-based model returns competitive results when benchmarked against a state-of-the-art time series generative model, a first impression suggests that the characteristics of the time series generated by WGAN-GP appear comparable to those obtained by the TimeGAN. To get a more precise answer and to make a grounded statement on the potential of the WGAN-GP and XIRP, we conduct the predictive and discriminative backtests described in Section \ref{sec:experiments}. 

Table \ref{tab:result_table} summarizes the scores for each model, metric, dataset, and representation with average scores on the left side, with the best result highlighted in boldface. On the right side of the table, we report the number of best scores. For example, for $S_D$ on the Brownian Motion dataset with its $n=10$ time series, the GASF performs best on 1/10, XIRP also on 1/10, naive on 8/10 and the time series representation of the TimeGAN performs best on 0/10 series. Note that we use IRC to convert the results from image-based models.
\begin{table*}[!htbp]
    \centering
    \resizebox{\linewidth}{!}{
    \begin{tabular}{ccccccc|cccc}
    \toprule
     Data & $n$ & Metric & TimeGAN & \multicolumn{3}{c}{WGAN} & \multicolumn{4}{|c}{\# Best Scores}\\
    & & & TS & GASF & XIRP & Naive & TS & GASF & XIRP & Naive\\
     \midrule
    Sine & 9 &  $S_D$ & 0.584 & \textbf{0.670} & 0.669 & 0.523 & 2 & 6 & 1 & 0\\
      & & & \footnotesize{(0.020)} & \footnotesize{(0.011)}& \footnotesize{(0.014)} & \footnotesize{(0.036)} & & & &\\
    Sine & 9 & $S_P$ & 0.469 & \textbf{0.079} & 0.079 & 0.4298 & 0 & 6 & 2 & 1\\
      & & & \footnotesize{(0.140)} & \footnotesize{(0.005)}& \footnotesize{(0.005)} & \footnotesize{(0.121)}& & & &\\
    Noisy Sine & 10 &  $S_D$ & 0.412 & 0.402 & \textbf{0.423}& 0.532  & 1 & 2 & 3 & 4\\
      & & & \footnotesize{(0.026)} & \footnotesize{(0.047)}& \footnotesize{(0.055)} & \footnotesize{(0.028)}& & & &\\
    Noisy Sine & 10 &  $S_P$ & 0.309 & \textbf{0.105} & \textbf{0.105}& 0.329 & 1 & 5 & 4 & 0\\
      & & & \footnotesize{(0.027)} & \footnotesize{(0.015)}& \footnotesize{(0.015)} & \footnotesize{(0.04)}& &\\
     Brownian Motion & 10 &  $S_D$ & 0.378 & 0.314 & 0.334 & \textbf{0.556} & 0 & 1 & 1 & 8\\
      & & & \footnotesize{(0.031)} & \footnotesize{(0.048)}& \footnotesize{(0.054)} & \footnotesize{(0.026)} & & \\
    Brownian Motion & 10 &  $S_P$ & 0.248 & \textbf{0.095} & 0.095 & 0.250   & 0 & 6 & 3 & 1 \\
      & & & \footnotesize{(0.03)} & \footnotesize{(0.01)}& \footnotesize{(0.011)} & \footnotesize{(0.037)}& & & & \\
     Merton Process & 10 &  $S_D$ & \textbf{0.462} & 0.456 & \textbf{0.462} & 0.416 & 5 & 1 & 3 & 1\\
      & & & \footnotesize{(0.019)} & \footnotesize{(0.047)}& \footnotesize{(0.052)} & \footnotesize{(0.029)} & &\\
    Merton Process & 10 &  $S_P$ & 0.270 & \textbf{0.102} & \textbf{0.102}& 0.278 & 1 & 4 & 5 & 0\\
      & & & \footnotesize{(0.006)} & \footnotesize{(0.006)}& \footnotesize{(0.006)} & \footnotesize{(0.017)} & & &\\
    Power Law & 9 &  $S_D$ & \textbf{0.670} & 0.441 & 0.442 & 0.625 & 8 & 0 & 0 & 1 \\
      & & & \footnotesize{(0.015)} & \footnotesize{(0.075)}& \footnotesize{(0.064)} & \footnotesize{(0.037)}& & &\\
    Power Law  & 9 &  $S_P$ & 0.377 & \textbf{0.125} & \textbf{0.125} & 0.360 & 1 & 5 & 3 & 0\\
      & & & \footnotesize{(0.105)} & \footnotesize{(0.093)}& \footnotesize{(0.093)} & \footnotesize{(0.106)} & & &\\
    Energy Data & 27 & $S_D$ & \textbf{0.591} & 0.450 & 0.445 & 0.590 & 3 & 4 & 4 & 16\\
      & & & \footnotesize{(0.026)} & \footnotesize{(0.053)}& \footnotesize{(0.056)} & \footnotesize{(0.026)} & & & &\\
    Energy Data & 27 &  $S_P$ & 0.352 & 0.092 & \textbf{0.092} & 0.347 & 0 & 15 & 11 & 1\\
      & & & \footnotesize{(0.015)} & \footnotesize{(0.015)}& \footnotesize{(0.019)} & \footnotesize{(0.074)} & & &\\
    Stock Data & 5 &  $S_D$ & \textbf{0.593} & 0.527 & 0.510 & 0.549 & 2 & 2 & 0 & 1\\
      & & & \footnotesize{(0.022)} & \footnotesize{(0.060)}& \footnotesize{(0.051)}& \footnotesize{(0.026)}& & &\\
    Stock Data  & 5 &  $S_P$ & 0.360 & 0.160 & \textbf{0.100} & 0.362 & 1 & 2 & 2 & 0\\
      & & & \footnotesize{(0.048)} & \footnotesize{(0.034)}& \footnotesize{(0.026)} & \footnotesize{(0.052)}& & &\\
    Air Quality  & 13 &  $S_D$ & 0.315 & 0.630 & 0.629 & \textbf{0.631} & 1 & 4 & 3 & 5\\
      & & & \footnotesize{(0.022)} &\footnotesize{(0.017)} & \footnotesize{(0.019)} & \footnotesize{(0.016)} & & &\\
    Air Quality  & 13 & $S_P$ & 0.299 & 0.248 & 0.248 & \textbf{0.247} & 4 & 2 & 2 & 5\\
      & & & \footnotesize{(0.048)} & \footnotesize{(0.106)}& \footnotesize{(0.106)} & \footnotesize{(0.106)} & & & & \\
    Bike Share  & 7 &  $S_D$ & 0.508 & \textbf{0.667} & 0.660  & 0.664 & 0 & 3 & 2 & 2\\
      & & & \footnotesize{(0.026)} & \footnotesize{(0.017)} & \footnotesize{(0.015)}& \footnotesize{(0.015)}& & & &\\
    Bike Share  & 7 &  $S_P$ & 0.730 & 0.663 & \textbf{0.656} & 0.663 & 3 & 0 & 2 & 2\\
      & & & (\footnotesize{0.008}) & \footnotesize{(0.009)} & \footnotesize{(0.009)}& \footnotesize{(0.106)} & & & &\\
    \end{tabular}
    }
      \parbox{13cm}{\caption{Benchmarking Results with mean predictive score $S_P$ (the lower the better) and \\ mean discriminative score $S_D$ (the higher the better)\label{tab:result_table}}}
\end{table*}
A first look at Table \ref{tab:result_table} reveals that all representations obtain the best scores on some of the data sets. Across the nine data sets, the TimeGAN achieves the best $S_D$ score in 4/9 cases but never achieves the best $S_D$ score. WGAN-GP obtains the best $S_P$ score on 5/9 data sets and the best $S_D$ score on 2/9 datasets using GASF. Using the XIRP representation, the WGAN-GP scores best on 5/9 $S_P$ and 3/9 $S_D$ scores. Even training on a naive representation achieves the best result in some occasions (e.g., lowest $S_P$ score on 1/9 and highest $S_D$ score on 2/9 dataset). Considering the individual time series level, the best-performing model regarding the average score usually has the highest number of wins for the dataset. A striking exception is $S_D$ on the energy data. Even though the average discriminative score for the TimeGAN ($S^{T}_D=0.591$) and the WGAN-GP trained on the naive representation ($S^{W}_D=0.590$) are close, the latter wins on 16/27 scores. This imbalance indicates that the naive model generally performs better but failed in a few examples to create sufficiently similar instances, which has a strong negative impact on the average. Also, for $S_P$ on the energy dataset, we see that GASF wins on 15/27 time series, obtaining a score of $S^{G}_P=0.092$ but is slightly outperformed by the XIRP with an average of $S^{X}_P=0.092$, which only wins 11/27 time series. We attribute this to the same causes. 
The WGAN-GP model regularly yields similar results for GASF and XIRP. The predictive score, e.g. for the Merton Process, Noisy Sine, and Power Law match up to the fourth decimal place, sharing the small values. In total, the image-based methods won 14/18 scores. Since the sizes of the data sets vary, a closer look at the individual time series level provides additional information. Here, for $S_D$, the time series representation wins 22/100 cases, while GASF scores slightly better with 23/100, and XIRP wins 17/100. Surprisingly, the naive approach outperforms all other approaches, winning 38/100 of the discriminative scores.\newline
Regarding $S_P$, performance differences among the representations are substantial. The WGAN-GP trained on $R^N$ and TimeGAN score similarly, winning only 10/100 and 11/100 time series, respectively. The XIRP scores better with 34/100 and outperforms GASF, winning 45/100 time series.\newline
It becomes evident that the representations and models perform differently concerning the scores. While the naive representation shows the best results for $S_D$, it performs worst, closely followed by TimeGAN, on $S_P$. Approximately, GASF and XIRP win twice as many time series on $S_D$ compared to $S_P$, opposing the performance of the TimeGAN. 
Regarding the cross-representation comparison, the tests do not identify a superior representation across all scores. It rather depends on the application of the model. If the goal is to find the most similar instances, choosing a WGAN-GP using a naive representation is advisable. If, on the other side, the goal is to synthesize data to augment the existing data for training, then the choice falls on a WGAN-GP using GASF or XIRP as a representation. However, in both cases, the simpler WGAN-GP model outperformed the TimeGAN model, delivering significantly better results with the naive approach on $S_D$ and GASF on $S_P$. Since our naive representation ensured the same level of information, we can state that the WGAN-GP is competitive with the TimeGAN model.
Now that the average metric score and the number of best scores are known, we evaluate the ranking of each model. We calculated a ranking for $S_D$ and $S_P$ on each time series and assigned ranks from 1 (best) to 4 (worst).
In case image-based representations consistently outperform the TimeGAN, they would take places 1,2 and 3 (average rank of 2) while TimeGAN scores fourth (average rank of 4). Comparing the average rank puts the performance of each model and representation into perspective. Regarding $S_D$, we find that the WGAN-GP model has an average rank of $\tilde{r(S^{W}_D)}=2.31$ while the TimeGAN has an average rank of $\tilde{r(S^{T}_D)}=2.67$. For $S_P$, however, the WGAN-GP model has an average rank across all time series of $\tilde{r}_{W}=3.15$ while the TimeGAN has an average rank of $\tilde{r}_{T}=2.17$. The ranks support the impression from Table \ref{tab:result_table2} indicating that the WGAN-GP trained on image representations has higher $S_P$ and lower $S_D$.
\begin{table*}
    \begin{subtable}{.5\linewidth}
      \centering
        \begin{tabular}{ccc}
            \toprule
            Rep & \# best $S_D$ &  \# best $S_P$ \\
            \midrule
            Time Series & 22 & 11 \\
            GASF & 23 & \textbf{45} \\
            XIRP & 17 & 34 \\
            Naive & \textbf{38} & 10 \\
            \midrule
            $\sum$ & 100 & 100 \\
             & & \\
            \toprule
            Model &  \# best $S_D$ &  \# best $S_P$ \\
            \midrule
            TimeGAN & 22 & 11 \\
            WGAN-GP & \textbf{78} & \textbf{89} \\
            & & \\
            \midrule
            $\sum$ & 100 & 100 \\
            & & \\
        \end{tabular}
        \caption{}
    \end{subtable}%
    \begin{subtable}{.5\linewidth}
      \centering
        \begin{tabular}{ccc}
            \toprule
            Mode/Rep. & $\overline{r}_{S_D}$  & $\overline{r}_{S_P}$\\
            \midrule
            GASF & 2.40 & 1.72 \\
            XIRP & 2.42 & \textbf{1.67} \\
            Naive & \textbf{2.11} & 3.06 \\
            TS & 2.67 & 3.15 \\
            \midrule
            TimeGAN & 2.67 & 3.15\\
            WGAN-GP & \textbf{2.31} & \textbf{2.17}\\
            & & \\
            & & \\
            & & \\
            & & \\
            & & \\ 
            & & \\
            & & \\
        \end{tabular}
        \caption{}
    \end{subtable}
     \caption{Evaluation of model and representation by (a) number of best scores and (b) average ranks from 1 (best) to 4}\label{tab:result_table2}
    
\end{table*}

A more granular look at the ranks for each representation in Table \ref{tab:result_table2} (b) shows that for $S_D$, the naive approach scores best and GASF, as well as XIRP, perform almost equally, which supports the impression obtained from Table \ref{tab:result_table}. Overall the WGAN-GP achieves a mean rank of 2.31, while the average rank of the synthesis by TimeGAN is 2.67. The lower rank makes the WGAN-GP the winner on the discriminative side. Regarding $S_P$, we see XIRP scores best, with an average rank of 1.67, closely followed by GASF. Here the difference in performance is significantly larger since the WGAN-GP with naive representation and WGAN-GP perform equally poorly. Thus, the average rank across representations is 2.17 for the WGAN-GP and 3.15 for the TimeGAN. According to this evaluation, the WGAN-GP model performs better across representations and models than the TimeGAN.

\begin{figure*}[htbp]
    \centering
    \subfloat[]{\includegraphics[width=0.45\textwidth,trim={1cm 1cm 1cm 1cm}]{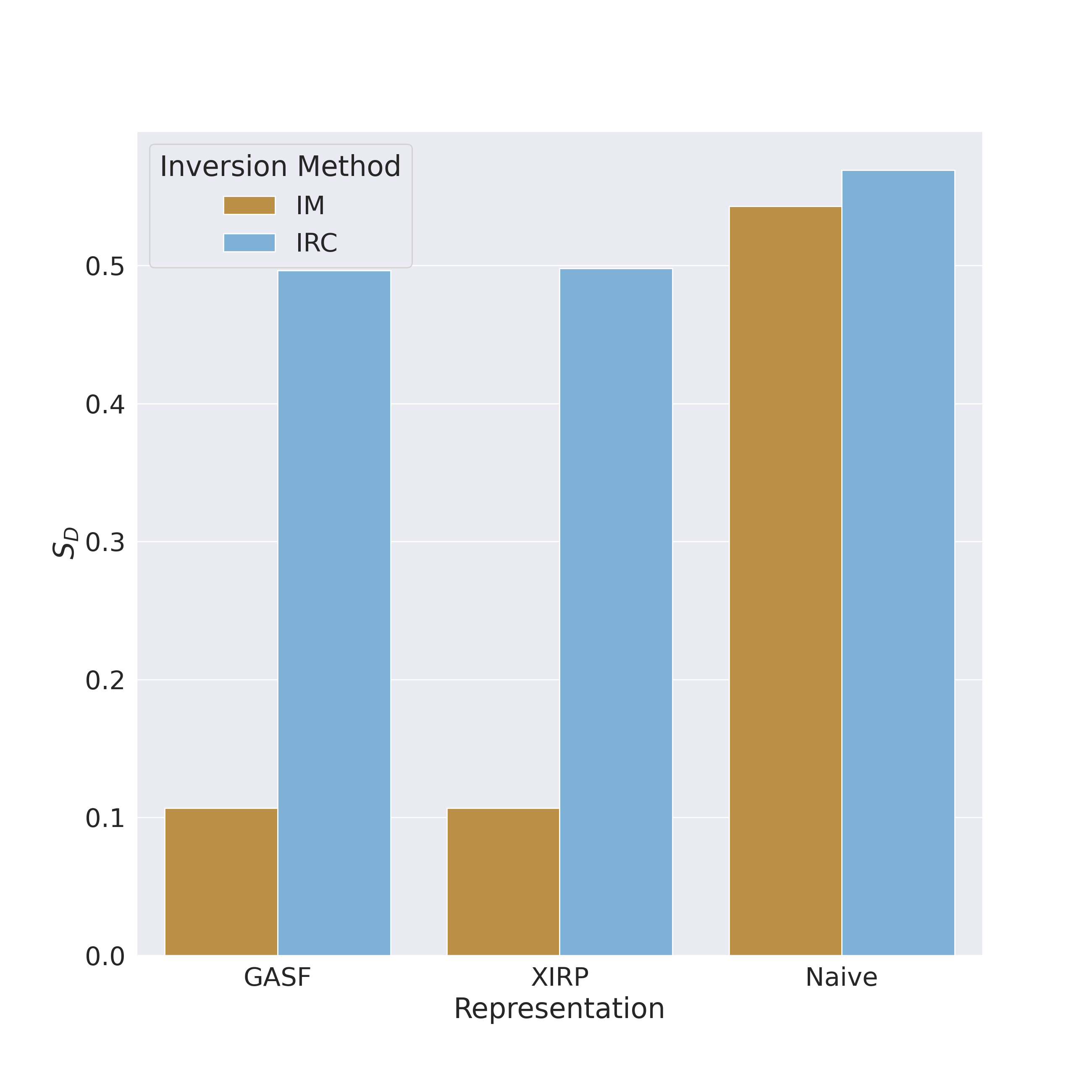}}
    \vspace{5pt}
    \hspace{.5cm}
    \subfloat[]{\includegraphics[width=0.45\textwidth,trim={1cm 1cm 1cm 1cm}]{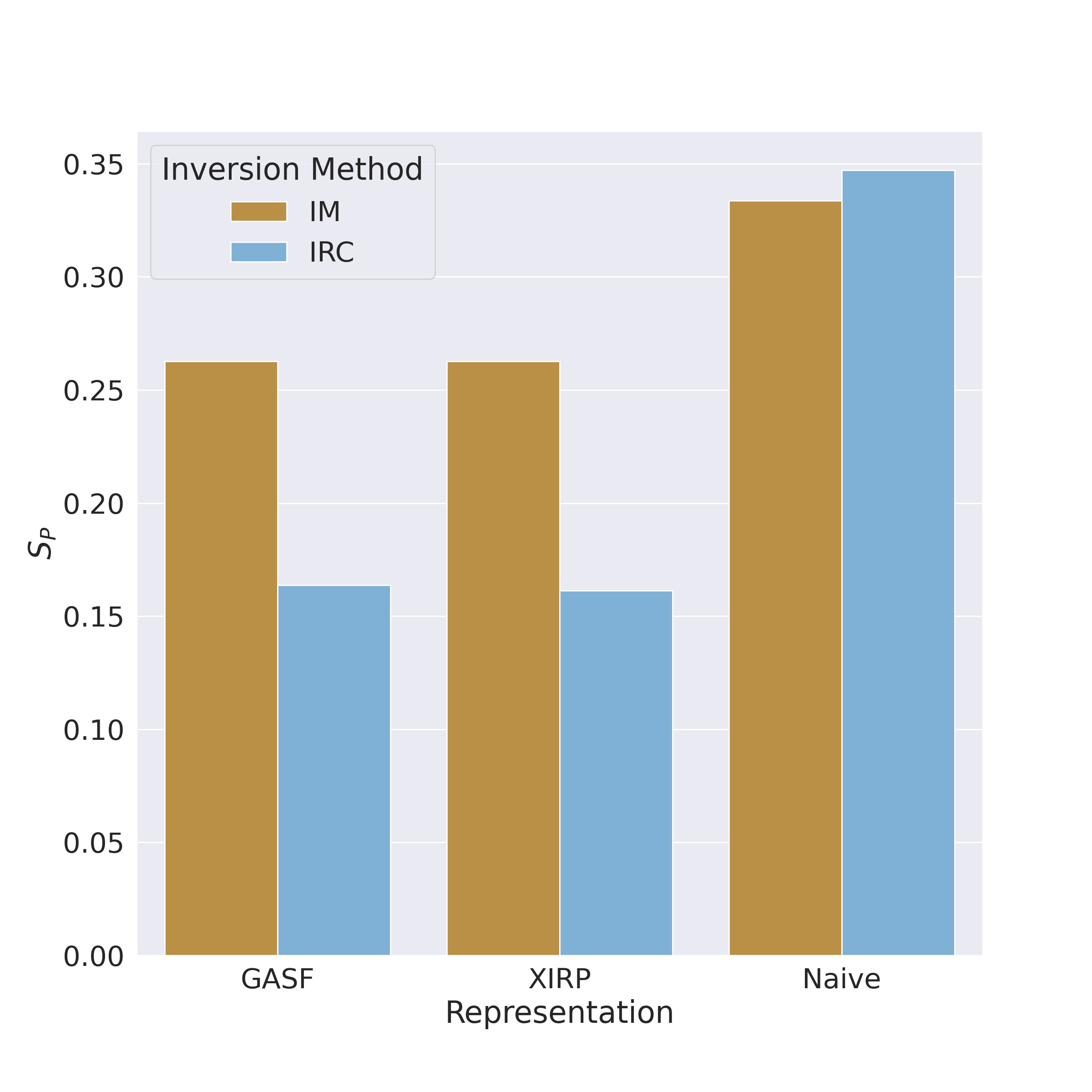}}
    \vspace{5pt}
    \caption{Metric scores by representation and inversion method for (a) $S_D$ and (b) $S_P$}
    \label{fig:improvements}
\end{figure*}

Tables \ref{tab:result_table} and \ref{tab:result_table2} have shown that the performance of WGAN-GP is at least competitive and, in many cases, superior to those of the TimeGAN model. The upper section introduces two inversion methods of which the ICR results from observing the harmful effects of IM on the training data. Visual inspections of synthetic time series obtained by IM and ICR have shown that IM returns smoother trajectories, which differ from the input time series and are thus easy to detect in the backtests. Figure \ref{fig:improvements} compares the performance for each representation and inversion method for each $S_D$ and $S_P$. A look at Figure \ref{fig:improvements} (a) shows that the IM inversion obtains a lower score on the discriminative metric. Since a high discriminative metric is best, we see that IRC improves the score for each representation. 
Improvements are most prominent in $S_D$, where both representations show gains from \raisebox{-0.8ex}{\~{}}4.8\% in the naive case to \raisebox{-0.8ex}{\~{}}360\% for GASF and XIRP. For the predictive score, the improvements range from -3.9\% for the naive representation, while XIRP and GASF improve by >60\%. Again, we see that XIRP and GASF benefit almost to the same extent from IRC while the naive model performs \raisebox{-0.8ex}{\~{}} 3.9\% worse, which is unexpected given the large increase for the other representations. Table \ref{fig:improvements} in the appendix summarises the computed effects.\newline
The introduction of the ICR has substantially improved performance regarding similarity and predictive ability, and the discovery of its superiority may also benefit other researchers in similar situations. Also, regarding computation times, we observe an advantage favoring the WGAN. Training on an Nvidia Tesla T4 GPU, it takes \raisebox{-0.8ex}{\~{}}10 minutes for a single time series, including the backtests compared to \raisebox{-0.8ex}{\~{}} 25 minutes for the TimeGAN model. The omnipresent trade-off in machine learning between model performance and computation time also applies here; given more training epochs and increased network size, it is likely also $S_D$ can be improved such that the slight difference in the average ranks of $S_D$ of  \raisebox{-0.8ex}{\~{}}5\% flips in favor of the WGAN.
Training hyperparameters such as batch size, epochs, and learning rates have remained at the default value proposed in the TimeGAN paper throughout all experiments. Thus, we have a well-grounded statement on competitive performance. However, we cannot give all credit to the new representation since comparing two inherently different neural networks always leaves room for other factors contributing to the success. \newline
Considering that the observed differences are substantial, we still conclude confidently that the proposed representation has contributed to the improvement in performance. In appreciating this result, it is essential to recall that the WGAN-GP model has a substantially easier structure with just one loss function and two networks compared to five loss functions and five networks in the TimeGAN model. \newline
The WGAN-GP model is a single generative model of a variety of models. In future research, an interesting approach would be to compare the results obtained with the WGAN-GP to other competitive approaches, such as diffusion models. A more thorough study of the statistical properties of the results could also answer the question under which conditions a successful synthesis is possible. Calculating statistics (mean, variance, skewness, entropy, etc.) for all input data sets and relating them to models and representations could provide insights under which circumstances it is unlikely that augmenting the data yields a benefit.
Further, it would be promising to follow the approach in the multivariate setting, similar to the one already used by TimeGAN, and to evaluate how well the models catch the relationships among the individual time series. Increasing the dimensionality also motivates an extension of research on multivariate versions of the recurrence plots to determine which form is the most efficient encoding for this case. Another exciting path would be evaluating different data augmentation steps for the WGAN-GP. Since the WGAN-GP relies on image representation, many common preprocessing steps could improve the generative performance, such as cropping, zooming, or shifting. Despite the success of the CNN model, it requires more  studies to explore the possibilities of using transformer models for the generator and discriminator using the same gradient penalty loss. Due to their attention mechanism, transformers can speed up training and exceed CNNs performance as the latter do not encode the relative position of different features or patterns in the time series. Choosing large filters to track long-range dependencies within a time series increases the representational capacity of the network, but doing so concedes a loss in computational and statistical efficiency. Even though the paper offers insight into the benefits of leveraging image-based generative adversarial networks for time series generation, there are plenty of potential research paths and aspects worth discovering.

\section{Conclusion}
The paper proposes a new two-dimensional image representation for time series, the Extended Intertemporal Return Plot (XIRP), to leverage the advances of image-based generative models for time series. 
It preserves the dynamics in a scale-invariant way without losing the ability to invert a time series numerically to its original scale, a significant advantage compared to other image representations. The XIRP can thus function as a powerful representation in regression and classification tasks.
Relying on a Wasserstein GAN with gradient penalty (WGAN-GP) for the generation process, we validated the XIRP on 100 time series from 9 data sets regarding similarity and predictive ability measures. The results show that an off-the-shelf generative model trained on image-based representations successfully generates synthetic time series while reducing model complexity and computation time. Backed by our experiments, the newly proposed XIRP captures the dynamics and characteristics of the time series. The approach outperforms other image representations like the Gramian Angular Summation Field (GASF) regarding predictive ability in a ranked performance comparison. In a cross-model comparison with an RNN-based TimeGAN model, the WGAN-GP trained on GASF or XIRP outperforms the benchmark regarding predictive ability. Surprisingly, the WGAN-GP trained on a naive representation of column-wise stacked time series performed best on the similarity test and outperformed GASF, XIRP, and TimeGAN. Our results indicate the absence of a generally superior representation across datasets and scores. Nonetheless, the proposed approach of synthesizing time series with an image-based model is generally superior. It outperforms the benchmark regarding sample quality and predictive ability for synthetic training data. The ranked performance and scores provided insights regarding the relative performance of models and representations. They showed that WGAN-GP trained on XIRP scores best regarding predictive ability, and a WGAN-GP trained on a naive representation scored best regarding similarity. Our approach shows that an efficient, invertible image representation such as XIRP can capture the dynamics and allows simple off-the-shelf generative models to generate time series with high predictive ability. Due to the simple representation and inversion, our approach enables the application of any image-based generative model for synthesizing time series data.
\newpage
\section{Appendix}
\onecolumn
\begin{landscape}
\renewcommand{\arraystretch}{2}
\begin{table}[htbp]
    \resizebox{\linewidth}{!}{\begin{tabular}{>{\centering\arraybackslash}p{5.0cm} |>{\centering\arraybackslash}p{2.4cm} >{\centering\arraybackslash}p{1.5cm} >{\centering\arraybackslash}p{2.4cm} >{\centering\arraybackslash}p{5cm} >{\centering\arraybackslash}p{1.7cm} >{\centering\arraybackslash}p{1.7cm} >{\centering\arraybackslash}p{2.4cm}}
    
& \textbf{Invertible} & \textbf{Scale-invariant} & \textbf{Encoding} & \textbf{Modeling Aspect} & \textbf{Classifi-cation} & \textbf{Regr-ession} & \textbf{Feature Scale} \\ \hline

\textbf{Binary Recurrence Plot} & no & no & deterministic & exceedances & yes & no & $x \in \mathbb{R}_{>0}$ \\

\textbf{Unthresholded Recurrence Plot} & yes, given $x_0$  & no & deterministic & difference in raw values & yes & yes & $x \in \mathbb{R}_{>0}$ \\

\textbf{Markov Transition Fields} & no & no & probabilistic & transition probabilities & yes & no & $x\in \mathbb{R}$ \\

\textbf{Gramian Angular Summation Field}  & yes  & no & deterministic & sum of polar coordinates & yes & yes & $0\leq x \leq 1$ \\

\textbf{Gramian Angular Difference Field} & yes, given $x_0$  & no & deterministic & difference of polar coordinates & yes & yes & $-1\leq x \leq 1$ \\

\textbf{Spectrograms} & no & no & probabilistic & frequencies of occurrence & yes & no & $x\in \mathbb{R}$ \\

\textbf{Waveform} & no (approximation is possible) & no & deterministic & shape of its graph as a function of time & yes & yes & $x\in \mathbb{R}$\\

\textbf{IRP} & yes, given $x_0$ & yes & deterministic & logarithmic percentage change & yes & yes & $x \in \mathbb{R}_{>0}$   \\

\textbf{XIRP} & yes, independent of $x_0$ & partially & deterministic & logarithmic percentage change and raw time series & yes & yes & $x \in \mathbb{R}_{>0}$ \\

\end{tabular}}
  \caption{Image Representations and Characteristics
  \label{tab:characteristics_table}}
\end{table}
\end{landscape}

\begin{table}[htbp]   
\centering
\begin{center}
    \begin{tabular}{ccccc}
    \toprule
    Rep.  &    $S$  &   $IM$ &  $IRC$  &  $\Delta$\% \\
    \midrule
    GASF & $S_D$ & 0.107 & 0.496 & 365.423 \\
    GASF & $S_P$ & 0.263 & 0.164 &  60.308 \\
    XIRP & $S_D$ &  0.107 & 0.498 & 366.619 \\
    XIRP & $S_P$ &  0.263 & 0.161  & 62.841 \\
    Naive & $S_D$ &  0.543 & 0.569  &  4.808 \\
    Naive & $S_P$ &  0.334 & 0.347 &  -3.881 \\
    & & \\
  \end{tabular}
   \caption{Score by Metric, Representation and Inversion Type}
  \label{tab:improvements}
\end{center}
\end{table}



\newpage
\bibliographystyle{apalike} 
\bibliography{references}

\begin{thebibliography}{}

\bibitem[Alcaraz and Strodthoff, 2022]{alcaraz_diffusion-based_2022}
Alcaraz, J. M.~L. and Strodthoff, N. (2022).
\newblock Diffusion-based time series imputation and forecasting with
  structured state space models.

\bibitem[Ali et~al., 2019]{ali_timecluster_2019}
Ali, M., Jones, M.~W., Xie, X., and Williams, M. (2019).
\newblock {TimeCluster}: dimension reduction applied to temporal data for
  visual analytics.
\newblock {\em The Visual Computer}, 35(6):1013--1026.

\bibitem[Allahyani et~al., 2023]{allahyani_divgan_2023}
Allahyani, M., Alsulami, R., Alwafi, T., Alafif, T., Ammar, H., Sabban, S., and
  Chen, X. (2023).
\newblock {DivGAN}: {A} diversity enforcing generative adversarial network for
  mode collapse reduction.
\newblock {\em Artificial Intelligence}, 317:103863.

\bibitem[Arjovsky et~al., 2017]{arjovsky_wasserstein_2017}
Arjovsky, M., Chintala, S., and Bottou, L. (2017).
\newblock Wasserstein {GAN}.
\newblock {\em arXiv:1701.07875 [cs, stat]}.
\newblock arXiv: 1701.07875.

\bibitem[Barra et~al., 2020]{barra_deep_2020}
Barra, S., Carta, S.~M., Corriga, A., Podda, A.~S., and Recupero, D.~R. (2020).
\newblock Deep learning and time series-to-image encoding for financial
  forecasting.
\newblock {\em IEEE/CAA Journal of Automatica Sinica}, 7(3):683--692.

\bibitem[Bengio, 2009]{bengio_learning_2009}
Bengio, Y. (2009).
\newblock Learning {Deep} {Architectures} for {AI}.
\newblock {\em Foundations and Trends® in Machine Learning}, 2(1):1--127.

\bibitem[Cao et~al., 2014]{gaussiantrees2014}
Cao, H., Tan, V., and Pang, J. (2014).
\newblock A parsimonious mixture of gaussian trees model for oversampling in
  imbalanced and multimodal time-series classification.
\newblock {\em IEEE transactions on neural networks and learning systems},
  25:2226--2239.

\bibitem[Carvajal-Patiño and Ramos-Pollán,
  2022]{carvajal-patino_synthetic_2022}
Carvajal-Patiño, D. and Ramos-Pollán, R. (2022).
\newblock Synthetic data generation with deep generative models to enhance
  predictive tasks in trading strategies.
\newblock {\em Research in International Business and Finance}, 62:101747.

\bibitem[Chen et~al., 2020]{chen_effectiveness_2020}
Chen, Y., Ji, A., Babajiyavar, P.~A., Ahmadzadeh, A., and Angryk, R.~A. (2020).
\newblock On the effectiveness of imaging of time series for flare forecasting
  problem.
\newblock In {\em 2020 IEEE International Conference on Big Data (Big Data)},
  pages 4184--4191.

\bibitem[Cui et~al., 2022]{cui_lightweight_2022}
Cui, J., Zhong, Q., Zheng, S., Peng, L., and Wen, J. (2022).
\newblock A lightweight model for bearing fault diagnosis based on gramian
  angular field and coordinate attention.
\newblock {\em Machines}, 10(4):282.

\bibitem[Dash et~al., 2020]{michalowski_medical_2020}
Dash, S., Yale, A., Guyon, I., and Bennett, K.~P. (2020).
\newblock Medical time-series data generation using generative adversarial
  networks.
\newblock {\em Artificial Intelligence in Medicine}, 12299:382--391.

\bibitem[{de Paula} et~al., 2023]{DEPAULA2023119096}
{de Paula}, P.~O., {da Silva Costa}, T.~B., {de Faissol Attux}, R.~R., and
  Fantinato, D.~G. (2023).
\newblock Classification of image encoded ssvep-based eeg signals using
  convolutional neural networks.
\newblock {\em Expert Systems with Applications}, 214:119096.

\bibitem[Dhariwal and Nichol, 2021]{NEURIPS2021_49ad23d1}
Dhariwal, P. and Nichol, A. (2021).
\newblock Diffusion models beat gans on image synthesis.
\newblock In Ranzato, M., Beygelzimer, A., Dauphin, Y., Liang, P., and Vaughan,
  J.~W., editors, {\em Advances in Neural Information Processing Systems},
  volume~34, pages 8780--8794. Curran Associates, Inc.

\bibitem[Donahue et~al., 2019]{donahue_adversarial_2019}
Donahue, C., McAuley, J., and Puckette, M. (2019).
\newblock Adversarial {Audio} {Synthesis}.
\newblock {\em arXiv:1802.04208 [cs]}.
\newblock arXiv: 1802.04208.

\bibitem[Drew, 2012]{drew2012}
Drew, C. (2012).
\newblock A survey of sequential monte carlo methods for economics and finance.
\newblock {\em Econometric Reviews}, 31(3):245--296.

\bibitem[Dua and Graff, 2017]{Dua2019}
Dua, D. and Graff, C. (2017).
\newblock {UCI} machine learning repository.

\bibitem[Durall et~al., 2023]{DURALL2023105377}
Durall, R., Ghanim, A., Fernandez, M., Ettrich, N., and Keuper, J. (2023).
\newblock Deep diffusion models for seismic processing.
\newblock {\em Computers \& Geosciences}, page 105377.

\bibitem[Eckmann et~al., 1987]{eckmann_recurrence_1987}
Eckmann, J.-P., Kamphorst, S.~O., and Ruelle, D. (1987).
\newblock Recurrence {Plots} of {Dynamical} {Systems}.
\newblock {\em Europhysics Letters (EPL)}, 4(9):973--977.

\bibitem[Elalem et~al., 2022]{ELALEM2022}
Elalem, Y.~K., Maier, S., and Seifert, R.~W. (2022).
\newblock A machine learning-based framework for forecasting sales of new
  products with short life cycles using deep neural networks.
\newblock {\em International Journal of Forecasting}.

\bibitem[Esteban et~al., 2017]{esteban_real-valued_2017}
Esteban, C., Hyland, S.~L., and Rätsch, G. (2017).
\newblock Real-valued ({Medical}) {Time} {Series} {Generation} with {Recurrent}
  {Conditional} {GANs}.
\newblock {\em arXiv:1706.02633 [cs, stat]}.
\newblock arXiv: 1706.02633.

\bibitem[Gillioz et~al., 2020]{gillioz_overview_2020}
Gillioz, A., Casas, J., Mugellini, E., and Khaled, O.~A. (2020).
\newblock Overview of the transformer-based models for nlp tasks.
\newblock In {\em 2020 15th Conference on Computer Science and Information
  Systems (FedCSIS)}, pages 179--183.

\bibitem[Goodfellow et~al., 2014]{goodfellow_generative_2014}
Goodfellow, I.~J., Pouget-Abadie, J., Mirza, M., Xu, B., Warde-Farley, D.,
  Ozair, S., Courville, A., and Bengio, Y. (2014).
\newblock Generative {Adversarial} {Networks}.
\newblock {\em arXiv:1406.2661 [cs, stat]}.
\newblock arXiv: 1406.2661.

\bibitem[Gulrajani et~al., 2017]{gulrajani_improved_2017}
Gulrajani, I., Ahmed, F., Arjovsky, M., Dumoulin, V., and Courville, A. (2017).
\newblock Improved {Training} of {Wasserstein} {GANs}.
\newblock {\em arXiv:1704.00028 [cs, stat]}.
\newblock arXiv: 1704.00028.

\bibitem[Herrera et~al., 2018]{HERRERA2018622}
Herrera, A.~M., Hu, L., and Pastor, D. (2018).
\newblock Forecasting crude oil price volatility.
\newblock {\em International Journal of Forecasting}, 34(4):622--635.

\bibitem[Iwanski and Bradley, 1998]{iwanski1998recurrence}
Iwanski, J.~S. and Bradley, E. (1998).
\newblock Recurrence plots of experimental data: To embed or not to embed?
\newblock {\em Chaos: An Interdisciplinary Journal of Nonlinear Science},
  8(4):861--871.

\bibitem[Karras et~al., 2019]{karras_style-based_2019}
Karras, T., Laine, S., and Aila, T. (2019).
\newblock A {Style}-{Based} {Generator} {Architecture} for {Generative}
  {Adversarial} {Networks}.
\newblock {\em arXiv:1812.04948 [cs, stat]}.
\newblock arXiv: 1812.04948.

\bibitem[Li et~al., 2020]{LI2020113680}
Li, X., Kang, Y., and Li, F. (2020).
\newblock Forecasting with time series imaging.
\newblock {\em Expert Systems with Applications}, 160:113680.

\bibitem[Li et~al., 2022]{michalowski_tts-gan_2022}
Li, X., Metsis, V., Wang, H., and Ngu, A. H.~H. (2022).
\newblock {TTS}-{GAN}: {A} {Transformer}-{Based} {Time}-{Series} {Generative}
  {Adversarial} {Network}.
\newblock In Michalowski, M., Abidi, S. S.~R., and Abidi, S., editors, {\em
  Artificial {Intelligence} in {Medicine}}, volume 13263, pages 133--143.
  Springer International Publishing, Cham.

\bibitem[Liang et~al., 2022]{LIANG2022117595}
Liang, M., Wu, S., Wang, X., and Chen, Q. (2022).
\newblock A stock time series forecasting approach incorporating candlestick
  patterns and sequence similarity.
\newblock {\em Expert Systems with Applications}, 205:117595.

\bibitem[Lipton et~al., 2017]{lipton_learning_2017}
Lipton, Z.~C., Kale, D.~C., Elkan, C., and Wetzel, R. (2017).
\newblock Learning to {Diagnose} with {LSTM} {Recurrent} {Neural} {Networks}.
\newblock arXiv:1511.03677 [cs].

\bibitem[Marwan et~al., 2007]{marwan_recurrence_2007}
Marwan, N., Carmenromano, M., Thiel, M., and Kurths, J. (2007).
\newblock Recurrence plots for the analysis of complex systems.
\newblock {\em Physics Reports}, 438(5-6):237--329.

\bibitem[McInnes et~al., 2020]{mcinnes_umap_2020}
McInnes, L., Healy, J., and Melville, J. (2020).
\newblock {UMAP}: {Uniform} {Manifold} {Approximation} and {Projection} for
  {Dimension} {Reduction}.
\newblock {\em arXiv:1802.03426 [cs, stat]}.
\newblock arXiv: 1802.03426.

\bibitem[Nagem et~al., 2018]{Nagem2018PredictingSF}
Nagem, T., Qahwaji, R., Ipson, S., Wang, Z., and Al-Waisy, A. (2018).
\newblock Deep learning technology for predicting solar flares from
  (geostationary operational environmental satellite) data.
\newblock {\em International Journal of Advanced Computer Science and
  Applications}, 9.

\bibitem[Nambiar et~al., 2020]{nambiar_transforming_2020}
Nambiar, A., Heflin, M., Liu, S., Maslov, S., Hopkins, M., and Ritz, A. (2020).
\newblock Transforming the {Language} of {Life}: {Transformer} {Neural}
  {Networks} for {Protein} {Prediction} {Tasks}.
\newblock In {\em Proceedings of the 11th {ACM} {International} {Conference} on
  {Bioinformatics}, {Computational} {Biology} and {Health} {Informatics}},
  pages 1--8, Virtual Event USA. ACM.

\bibitem[Naveed et~al., 2022]{naveed2022assessing}
Naveed, M.~H., Hashmi, U.~S., Tajved, N., Sultan, N., and Imran, A. (2022).
\newblock Assessing deep generative models on time series network data.
\newblock {\em IEEE Access}, 10:64601--64617.

\bibitem[Park et~al., 2022]{park2022neural}
Park, S.~W., Lee, K., and Kwon, J. (2022).
\newblock Neural markov controlled {SDE}: Stochastic optimization for
  continuous-time data.
\newblock In {\em International Conference on Learning Representations}.

\bibitem[Ramesh et~al., 2021]{dalle2021}
Ramesh, A., Pavlov, M., Goh, G., Gray, S., Voss, C., Radford, A., Chen, M., and
  Sutskever, I. (2021).
\newblock Zero-shot text-to-image generation.

\bibitem[Sarkar et~al., 2020]{sarkar2020neural}
Sarkar, A., Raj, A.~S., and Iyengar, R.~S. (2020).
\newblock Neural data augmentation techniques for time series data and its
  benefits.
\newblock In {\em 2020 19th IEEE International Conference on Machine Learning
  and Applications (ICMLA)}, pages 107--114. IEEE.

\bibitem[Scharth and Medeiros, 2009]{SCHARTH2009304}
Scharth, M. and Medeiros, M.~C. (2009).
\newblock Asymmetric effects and long memory in the volatility of dow jones
  stocks.
\newblock {\em International Journal of Forecasting}, 25(2):304--327.
\newblock Forecasting Returns and Risk in Financial Markets using Linear and
  Nonlinear Models.

\bibitem[Schäfer and Guhr, 2010]{SCHAFER20103856}
Schäfer, R. and Guhr, T. (2010).
\newblock Local normalization: Uncovering correlations in non-stationary
  financial time series.
\newblock {\em Physica A: Statistical Mechanics and its Applications},
  389(18):3856--3865.

\bibitem[Shahverdy et~al., 2020]{SHAHVERDY2020113240}
Shahverdy, M., Fathy, M., Berangi, R., and Sabokrou, M. (2020).
\newblock Driver behavior detection and classification using deep convolutional
  neural networks.
\newblock {\em Expert Systems with Applications}, 149:113240.

\bibitem[Sutcliffe, 1994]{sutcliffe_time-series_1994}
Sutcliffe, A. (1994).
\newblock Time-series forecasting using fractional differencing.
\newblock {\em Journal of Forecasting}, 13(4):383--393.

\bibitem[Tashiro et~al., 2021]{tashiro2021}
Tashiro, Y., Song, J., Song, Y., and Ermon, S. (2021).
\newblock Csdi: Conditional score-based diffusion models for probabilistic time
  series imputation.
\newblock In Ranzato, M., Beygelzimer, A., Dauphin, Y., Liang, P., and Vaughan,
  J.~W., editors, {\em Advances in Neural Information Processing Systems},
  volume~34, pages 24804--24816. Curran Associates, Inc.

\bibitem[Taylor, 2019]{TAYLOR20191193}
Taylor, N. (2019).
\newblock Forecasting returns in the vix futures market.
\newblock {\em International Journal of Forecasting}, 35(4):1193--1210.

\bibitem[Tran and Kukal, 2022]{TRAN2022102574}
Tran, Q.~V. and Kukal, J. (2022).
\newblock A novel heavy tail distribution of logarithmic returns of
  cryptocurrencies.
\newblock {\em Finance Research Letters}, 47:102574.

\bibitem[van~den Oord et~al., 2016]{oord2016wavenet}
van~den Oord, A., Dieleman, S., Zen, H., Simonyan, K., Vinyals, O., Graves, A.,
  Kalchbrenner, N., Senior, A., and Kavukcuoglu, K. (2016).
\newblock Wavenet: A generative model for raw audio.

\bibitem[Wang et~al., 2022]{wang_quantitative_2022}
Wang, Q., Pian, F., Wang, M., Song, S., Li, Z., Shan, P., and Ma, Z. (2022).
\newblock Quantitative analysis of raman spectra for glucose concentration in
  human blood using gramian angular field and convolutional neural network.
\newblock {\em Spectrochimica Acta Part A: Molecular and Biomolecular
  Spectroscopy}, 275:121189.

\bibitem[Wang and Oates, 2015]{wang2015imaging}
Wang, Z. and Oates, T. (2015).
\newblock Imaging time-series to improve classification and imputation.
\newblock {\em Twenty-Fourth International Joint Conference on Artificial
  Intelligence}.

\bibitem[Wen et~al., 2021]{Wen_2021}
Wen, Q., Sun, L., Yang, F., Song, X., Gao, J., Wang, X., and Xu, H. (2021).
\newblock Time series data augmentation for deep learning: A survey.
\newblock In {\em Proceedings of the Thirtieth International Joint Conference
  on Artificial Intelligence}. International Joint Conferences on Artificial
  Intelligence Organization.

\bibitem[Wilmott, 2006]{wilmott_paul_2006}
Wilmott, P. (2006).
\newblock {\em Paul {Wilmott} on quantitative finance}.
\newblock Wiley, Chichester Weinheim, 2. editon, reprinted edition.

\bibitem[Wu et~al., 2022]{wu_prediction_2022}
Wu, J.-L., Tang, X.-R., and Hsu, C.-H. (2022).
\newblock A prediction model of stock market trading actions using generative
  adversarial network and piecewise linear representation approaches.
\newblock {\em Soft Computing}.

\bibitem[Wyse, 2017]{wyse_audio_2017}
Wyse, L. (2017).
\newblock Audio {Spectrogram} {Representations} for {Processing} with
  {Convolutional} {Neural} {Networks}.
\newblock {\em arXiv:1706.09559 [cs]}.
\newblock arXiv: 1706.09559.

\bibitem[Yoon et~al., 2019]{yoon2019}
Yoon, J., Jarrett, D., and van~der Schaar, M. (2019).
\newblock Time-series generative adversarial networks.
\newblock In Wallach, H., Larochelle, H., Beygelzimer, A., d\textquotesingle
  Alch\'{e}-Buc, F., Fox, E., and Garnett, R., editors, {\em Advances in Neural
  Information Processing Systems}, volume~32. Curran Associates, Inc.

\bibitem[Zhang and Guo, 2020]{zhang_novel_2020}
Zhang, G. and Guo, J. (2020).
\newblock A novel ensemble method for hourly residential electricity
  consumption forecasting by imaging time series.
\newblock {\em Energy}, 203:117858.

\bibitem[Zhang et~al., 2023]{zhang_solargan_2023}
Zhang, Y., Schlueter, A., and Waibel, C. (2023).
\newblock {SolarGAN}: {Synthetic} annual solar irradiance time series on urban
  building facades via {Deep} {Generative} {Networks}.
\newblock {\em Energy and AI}, 12:100223.

\bibitem[Zhang et~al., 2022]{zhang_data_2022}
Zhang, Y., Zhou, Z., Liu, J., and Yuan, J. (2022).
\newblock Data augmentation for improving heating load prediction of heating
  substation based on {TimeGAN}.
\newblock {\em Energy}, 260:124919.

\bibitem[Zhu et~al., 2022]{ZHU2022115992}
Zhu, K., Zhang, S., Li, J., Zhou, D., Dai, H., and Hu, Z. (2022).
\newblock Spatiotemporal multi-graph convolutional networks with synthetic data
  for traffic volume forecasting.
\newblock {\em Expert Systems with Applications}, 187:115992.

\end{thebibliography}


\end{document}